\documentclass[10pt,twocolumn,letterpaper]{article}
\usepackage[accsupp]{axessibility} % Improves PDF readability for those with disabilities.
\usepackage{iccv}
\usepackage{times}
\usepackage{epsfig}
\usepackage{graphicx}
\usepackage{amsmath,bm}
\usepackage{amssymb}
\usepackage{multirow}
\usepackage{multicol}
\usepackage{tabularx,booktabs}
\usepackage{subcaption}
\usepackage[font=small]{caption}
\usepackage{hhline}
\usepackage[T1]{fontenc}
\usepackage{cellspace}
\usepackage{adjustbox}
\usepackage{float}
\usepackage{amsmath,amsfonts,bm}

\def\eqref#1{equation~\ref{#1}}
\def\1{\bm{1}}

\DeclareMathAlphabet{\mathsfit}{\encodingdefault}{\sfdefault}{m}{sl}
\SetMathAlphabet{\mathsfit}{bold}{\encodingdefault}{\sfdefault}{bx}{n}

\newcommand{\E}{\mathbb{E}}

\usepackage{capt-of,etoolbox}

\newcommand{\norm}[1]{\left\lVert#1\right\rVert}

\makeatletter
\newcommand\footnoteref[1]{\protected@xdef\@thefnmark{\ref{#1}}\@footnotemark}
\makeatother

\def\E{{\rm E}}
\def\N{{\mathcal N}}

\def\I{{\bf I}}

\def\x{{\bf x}}
\def\y{{\bf y}}
\def\z{{\bf z}}

\usepackage{color,xcolor}
\definecolor{MyDarkBlue}{rgb}{0,0.08,1}
\definecolor{MyDarkGreen}{rgb}{0.02,0.6,0.02}
\definecolor{MyDarkRed}{rgb}{0.8,0.02,0.02}
\definecolor{MyDarkOrange}{rgb}{0.40,0.2,0.02}
\definecolor{MyPurple}{RGB}{111,0,255}
\definecolor{MyRed}{rgb}{1.0,0.0,0.0}
\definecolor{MyGold}{rgb}{0.75,0.6,0.12}
\definecolor{MyDarkgray}{rgb}{0.66, 0.66, 0.66}

\usepackage[pagebackref=true,breaklinks=true,letterpaper=true,colorlinks,bookmarks=false]{hyperref}

\iccvfinalcopy % *** Uncomment this line for the final submission

\def\iccvPaperID{9265} % *** Enter the ICCV Paper ID here

\ificcvfinal\fi

\newcommand{\model}{PVD\xspace}
\newcommand{\myparagraph}[1]{\vspace{-10pt}\paragraph{#1}}
\newcommand{\sect}[1]{Section~\ref{#1}}

\newcommand{\fig}[1]{Figure~\ref{#1}}

\begin{document}

%%%%%%%%% TITLE
\title{3D Shape Generation and Completion through Point-Voxel Diffusion}

\author{Linqi Zhou\\
Stanford University
\and
Yilun Du\\
MIT
\and
Jiajun Wu\\
Stanford University
}

%\maketitle
\ificcvfinal\thispagestyle{empty}\fi

\twocolumn[
\maketitle
\vspace{-1.5em}
\centering
\includegraphics[width=\linewidth]{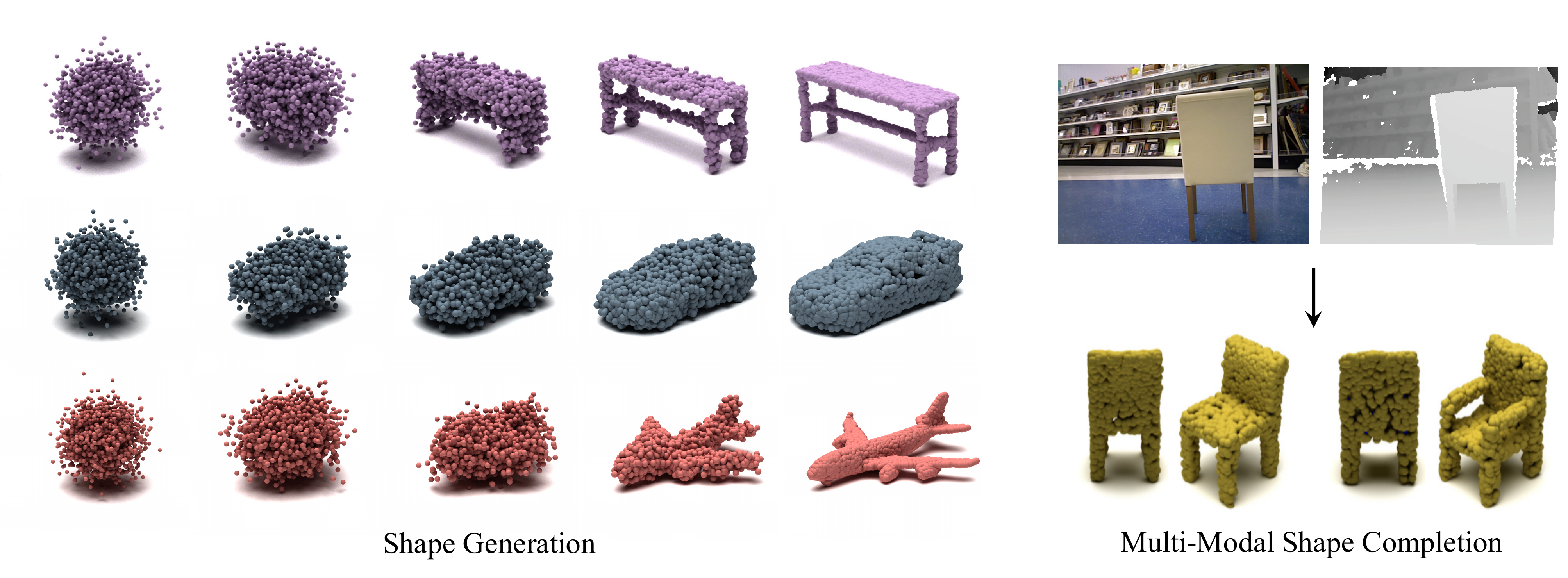}
\vspace{-20pt}
\captionof{figure}{The proposed Point-Voxel Diffusion (\model) is a new framework for generative modeling of 3D shapes. Left: tables, cars, and planes generated by our \model. It learns to sample from a Gaussian prior and to progressively remove noise to obtain sharp shapes. Right: two possible shapes completed from a real RGB-D image, each visualized in input and canonical views.}
\vspace{25pt}
\label{fig:teaser}

%\medbreak
%\vspace{.2em}
]

%%%%%%%%% ABSTRACT
\begin{abstract}
 %\vspace{-5pt}
   We propose a novel approach for probabilistic generative modeling of 3D shapes. Unlike most existing models that learn to deterministically translate a latent vector to a shape, our model, Point-Voxel Diffusion (\model), is a unified, probabilistic formulation for unconditional shape generation and conditional, multi-modal shape completion. \model marries denoising diffusion models with the hybrid, point-voxel representation of 3D shapes. It can be viewed as a series of denoising steps, reversing the diffusion process from observed point cloud data to Gaussian noise, and is trained by optimizing a variational lower bound to the (conditional) likelihood function. Experiments demonstrate that \model is capable of synthesizing high-fidelity shapes, completing partial point clouds, and generating multiple completion results from single-view depth scans of real objects.  \let\thefootnote\relax\footnotetext{Project page at \url{https://alexzhou907.github.io/pvd}}
   \vspace{-5pt}
\end{abstract}

%%%%%%%%% BODY TEXT

\section{Introduction}

%Three dimensional representations of the world are ubiquitous across domains in autonomous vehicles, visual arts, and robotics. In autonomous vehicles, incoming depth maps are aggregated into volumetric occupancy fields to enable effective navigation and collision avoidance. In creative arts, artists have access to 3D representations of shape, lighting, and material to enable convenient digital manipulation. In robotics, accurate models of object shape and material enable efficient robotic manipulation. Across these settings, the ability to do shape generation, completion, and deformation are highly important. 

Generative modeling of 3D shapes has extensive applications across vision, graphics, and robotics. %In autonomous vehicles, inferring the invisible based on partial observations (\eg, depth maps) enables effective navigation and collision avoidance. Artists may leverage generative models of shape, lighting, and material for content creation. In robotics, accurate modeling of object shape and material also facilitates robotic manipulation. Across these settings, the ability to generate, complete, and deform shapes is highly important. 
To perform well in these downstream applications, a good 3D generative models should be \emph{faithful} and \emph{probabilistic}. A faithful model generates shapes that are realistic to humans and, in cases where conditional inputs such as depth maps are available, respects such partial observations. A probabilistic model captures the under-determined, multi-modal nature of the generation and completion problem: it may sample and produce diverse shapes from scratch or from partial observations. As shown in \fig{fig:teaser}, when only the back of a chair is visible, good generative models should be able to produce multiple possible completed chairs, including those with arms and those without.

Existing shape generation models can be roughly divided into two categories. The first operates on 3D voxels~\cite{wu2016learning, huang20193d, xie2018learning, brock2016generative}, a natural extension of 2D pixels. While being straightforward to use, voxels demand prohibitively large memory when scaled to high dimensions, and are thus unlikely to produce results with high fidelity. The second class of models studies point cloud generation~\cite{achlioptas2018learning, gadelha2018multiresolution, zamorski2020adversarial, yang2019pointflow, kim2020softflow, klokov2020discrete} and has produced promising results. While being more faithful, these approaches typically view point cloud generation as a point generation process conditioned on shape encoding, which is obtained by deterministic encoders. When performing shape completion, these approaches are therefore unable to capture the multi-modal nature of the completion problem.

Recently, a new class of generative models, named probabilistic diffusion models, have achieved impressive performance on 2D image generation~\cite{sohl2015deep, ho2020denoising, song2020denoising}. These approaches learn a probabilistic model over a denoising process. Diffusion is supervised to gradually denoise a Gaussian noise to a target output, such as an image. Methods along this line, such as DDPM~\cite{ho2020denoising}, are inherently probabilistic and produce highly realistic 2D images.

Extending diffusion models to 3D is, however, technically highly nontrivial: a direct application of diffusion models on either voxel and point representation results in poor generation quality. This is because, first, pure voxels are binary and therefore not suitable for the probabilistic nature of diffusion models; second, point clouds demand permutation-invariance, which imposes infeasible constraints on the model. Experiments in \sect{sect:shape_gen} also verifies that a straightforward extension does not lead to reasonable results.

We propose Point-Voxel Diffusion (\model), a probabilistic and flexible shape generation model that tackles the above challenges by marrying denoising diffusion models with the hybrid, point-voxel representation of 3D shapes~\cite{liu2019point}. A point-voxel representation builds structured locality into point cloud processing; integrated with denoising diffusion models, \model suggests a novel, probabilistic way to generate high-quality shapes by denoising a Gaussian noise and to produce multiple completion results from a partial observation, as shown in \fig{fig:teaser}.

A unique strength of \model is that it is a unified, probabilistic formulation for unconditional shape generation and conditional, multi-modal shape completion. While multi-modal shape completion is a highly desirable feature in applications such as digital design or robotics, %it is challenging to learn due to difficulty in outputting realistic shapes while satisfying partial constraints. Consequently, 
past works on shape generation primarily use deterministic shape encoders and decoders to output a single possible completion in voxels or a point cloud. In contrast, \model can perform both unconditional shape generation and conditional shape completion in an integrated framework, requiring only minimal modifications to the training objective. It is thus capable of sampling multiple completion results depending on diffusion initialization. 

Experiments demonstrate that \model is capable of synthesizing high-fidelity shapes, outperforming multiple state-of-the-art methods. \model also delivers high-quality results on multi-modal shape completion from partial observations such as a partial point cloud or a depth map. In particular, we show that \model does well on multi-modal completion on multiple synthetic and real datasets, including ShapeNet~\cite{chang2015shapenet}, PartNet~\cite{Mo_2019_CVPR}, and single-view depth scans of real objects in the Redwood dataset~\cite{choi2016large}.

% Overall, our contribution are three-fold:
% \begin{itemize}
% %   \item We introduce denoising diffusion models to 3D shape synthesis and completion.
%   \item We present \model, a denoising diffusion model that obtains state-of-the-art 3D point cloud modeling.
%   \item We analyze different intermediate representations for \model generation, and show that the point-voxel representation are superior to either utilizing voxel or pointcloud representations.
%   \item With no architectural change, we further show \model obtains superior performance on shape completion, and can generate multiple diverse completions from single view depth scan, including real world depth scans.
% \end{itemize}

\section{Related Works}

\paragraph{Point cloud generative models.} Many prior works have explored point cloud generation in terms of auto-encoding~\cite{achlioptas2018learning, gadelha2018multiresolution, yang2018foldingnet}, single-view reconstruction~\cite{groueix2018papier,fan2017point,kurenkov2018deformnet,klokov2020discrete}, and adversarial generation~\cite{shu20193d,zamorski2020adversarial, valsesia2018learning}. Many of them rely on directly optimizing heuristic loss functions such as Chamfer Distance (CD) and Earth Mover's Distance (EMD), which are also used to evaluate generative quality. 

Some recent works take a different approach, viewing the 3D point clouds in light of probabilistic distributions. For example, Sun~\etal~\cite{sun2020pointgrow} view the point clouds from a probabilistic perspective and introduce autoregressive generation, but doing so requires ordering of the point clouds. GAN-based models and flow-based models~\cite{li2018point, achlioptas2018learning, yang2019pointflow, kim2020softflow, klokov2020discrete} also adopt a probabilistic view but separate shape-level distribution from point-level distribution. Among these models, PointFlow applies normalizing flow~\cite{papamakarios2019normalizing} to 3D point clouds, and Discrete PointFlow follows up using discrete normalizing flow with affine coupling layers~\cite{dinh2016density}. Shape Gradient Fields~\cite{cai2020learning}, unlike flow-based works, directly learn a gradient field that samples point clouds using Langevin dynamics. Our model is different from these models in that we do not distinguish point and shape distributions, and that we directly generate entire shapes starting from random noise.

\myparagraph{Point-voxel representation.} 3D shapes were conventionally rasterized into voxel grids and processed using 3D convolution~\cite{choy20163d, wang2019voxsegnet}. Due to the correspondence between voxels and 2D pixels, many works have explored voxel-based classification and segmentation using volumetric convolution~\cite{maturana2015voxnet, qi2016volumetric, le2018pointgrid, tatarchenko2017octree, wang2019voxsegnet, cciccek20163d}. Voxel-based generative models have similarly proven successful~\cite{wu2016learning, huang20193d, xie2018learning}. However, voxel grids are memory-intensive and they grow cubically with increase in dimension, so they cannot be scaled to a high resolution. 

Point clouds, on the other hand, are detailed samples from smooth surfaces and do not suffer from the grid effect of usually low-resolution voxels and do not require as much memory for processing. Researchers have explored point cloud classification and segmentation~\cite{qi2017pointnet, qi2017pointnet, wang2019dynamic} and most assume point cloud processing networks are permutation-invariant. Permutation-invariance is a strong condition to be imposed on the architecture and we empirically find that direct extension of 2D methods to either permutation-invariant point clouds or voxels do not work well. We therefore explore a separate point-voxel representation~\cite{li2018pointcnn, shi2020pv}, and our work is most related to point-voxel CNN~\cite{liu2019point}, which proposes to voxelize the point clouds for 3D convolution. We use it as the backbone of our generative model due to its exploitation of the strong spacial correlation inherent in point cloud data.

\myparagraph{Energy-based models and denoising diffusion models.}
Energy-based models (EBMs) and denoising diffusion models are two classes of generative models that formulate generation as an iterative refinement procedure. Energy based models~\cite{lecun2006tutorial, du2019implicit, nijkamp2019learning} learn an energy landscape over input data, where local minima correspond to high-fidelity samples, which are obtained by Langevin dynamics~\cite{du2019implicit, nijkamp2019learning}. In contrast, denoising diffusion models~\cite{sohl2015deep, ho2020denoising, song2020denoising} learn a probabilistic model over a denoising process on inputs. Diffusion is supervised to gradually denoise a Gaussian noise to a target output. This form of supervision can be seen as supervision of the gradient of a log probability distribution~\cite{song2020denoising} as in score matching EBM~\cite{hyvarinen2005estimation,swersky2011autoencoders}. Our work builds on these related existing approaches and we explore the 3D domain, which is challenging and fundamentally different from 2D images. A concurrent work on point cloud diffusion model~\cite{luo2021diffusion} views point cloud generation as a \textit{conditional} generation problem and uses an additional encoder for shape latents. Ours, however, adopts an \textit{unconditional} approach, ridding the need for additional shape encoders, and uses a different hybrid, point-voxel representation for processing shapes. In addition to generating high-quality 3D shapes, we also show that our model can be modified with no architectural change to perform on conditional generation tasks such as shape completion. We also demonstrate its effectiveness on real-world scans. 
\begin{figure}[t]
    \centering
    \includegraphics[width=\linewidth]{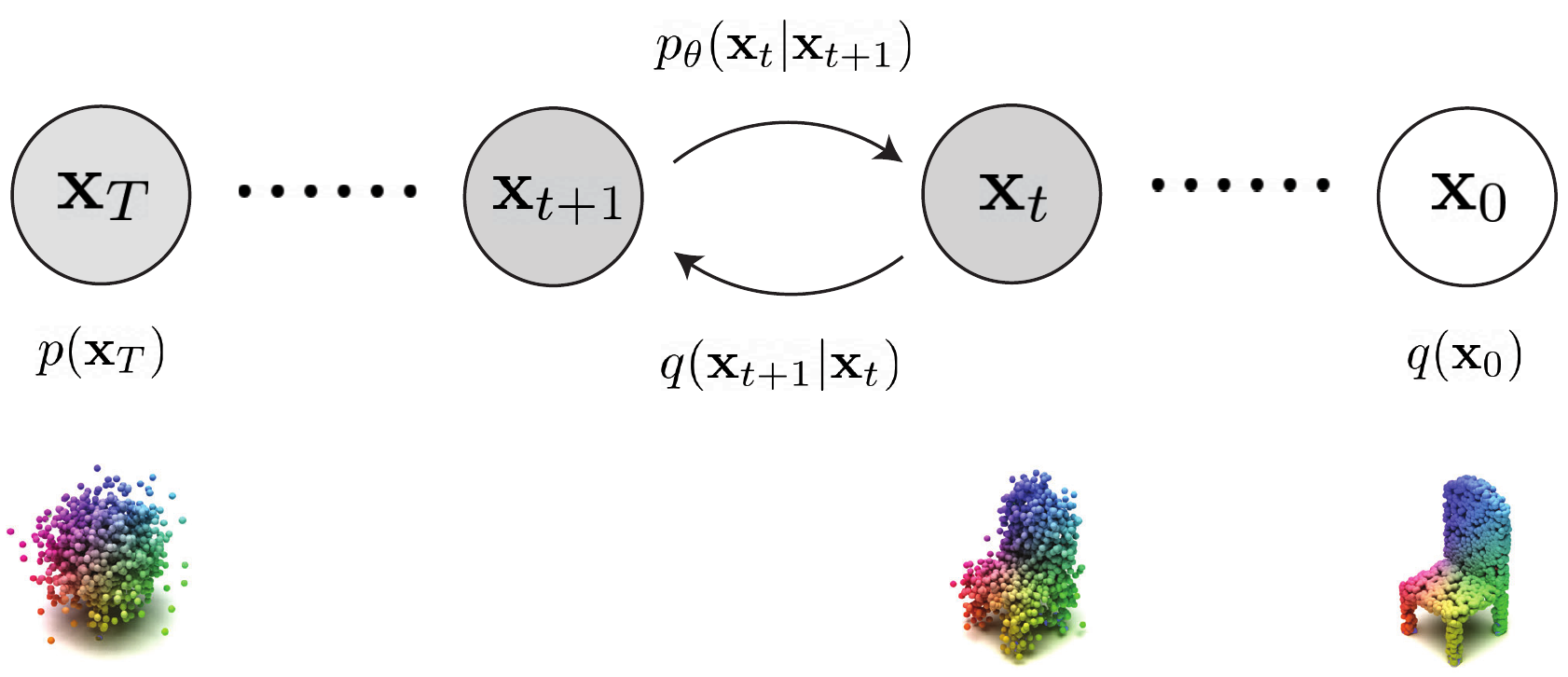}
    \caption{Visualization of the diffusion and generative process. To generate, Gaussian noise is sampled from $p(\x_T)$ and noise is progressively removed by $p_\theta(\x_t|\x_{t+1})$. Symmetrically, the diffusion process gradually adds noise by $q(\x_{t+1}|\x_t)$. We utilize a closed-form expression for each $q(\x_{t+1}|\x_t)$, allowing  $p_\theta(\x_t|\x_{t+1})$ to be learned by simply matching the posterior $q(\x_t|\x_{t+1},\x_0)$ of the corresponding forward transition probability.}
    \label{fig:demo}
\end{figure}

\section{Point-Voxel Diffusion}
In this section we introduce Point-Voxel Diffusion (\model), a denoising diffusion probabilistic model for 3D point clouds. We start by describing our formulation, followed by the training objective for shape generation, and end with the modified objective we proposed for shape completion from partial observation. For all our discussions below, we assume each of our data points are a set of $N$ points with $xyz$-coordinates and is denoted as $\x \in \mathbb{R}^{N\times 3}$. Our model is parameterized as a single point-voxel CNN~\cite{liu2019point}. 

\subsection{Formulation}
The denoising diffusion probabilistic model is a generative model where generation is modeled as a denoising process. Starting from Gaussian noise, denoising is performed until a  sharp shape is formed.  In particular, the denoising process produces a series of shape variables with decreasing levels of noise, denoted as $\x_T, \x_{T-1}, ..., \x_0$, where $\x_T$ is sampled from a Gaussian prior and $\x_0$ is the final output. 

To learn our generative model, we define a ground truth diffusion distribution $q(\x_{0:T})$ (defined by gradually adding Gaussian noise to the ground truth shape), and learn a diffusion model  $p_\theta(\x_{0:T})$, which aims to invert the noise corruption process. We factor both probability distributions into products of Markov transition probabilities:
\begin{equation}
\label{eq:factorize}
\begin{aligned}
    q(\x_{0:T}) &= q(\x_0)\prod_{t=1}^{T} q(\x_t | \x_{t-1}),\\
p_\theta(\x_{0:T}) &=  p(\x_T)\prod_{t=1}^{T} p_\theta(\x_{t-1} | \x_t),
\end{aligned}
\end{equation}
where $q(\x_0)$ is the data distribution and $p(\x_T)$ is a standard Gaussian prior. Here, $q(\x_t | \x_{t-1})$ is named the \textit{forward process}, diffusing data into noise; accordingly, $q(\x_{t-1}|\x_t)$ is named the \textit{reverse process}. $p_\theta(\x_{t-1} | \x_t)$ is named the \textit{generative process}, which we learn, that generates realistic samples by approximates the reverse process. To enable closed-form evaluation, the transition probabilities are also parameterized as Gaussian distributions. We illustrate the processes in Figure~\ref{fig:demo}. Given a pre-determined increasing sequence of Gaussian noise values $\beta_1, ...,  \beta_T$,\footnote{We leave derivation and implementation details to Appendix.\label{details}} each transition probability can be defined as
\begin{equation}\label{eq:transition}
\begin{aligned}
q(\x_t | \x_{t-1}) &:= \N(\sqrt{1-\beta_t} \x_{t-1}, \beta_t \I),    \\
p_\theta( \x_{t-1}|\x_t) &:= \N(\mu_\theta(\x_t, t), \sigma_t^2  \I).
\end{aligned}
\end{equation},
where $\mu_{\theta}(\x_t, t)$ represents the predicted shape from our generative model at timestep $t-1$. 
Empirically, we found that setting $\sigma_t^2 = \beta_t$ works well. Intuitively, the forward process can be seen as gradually injecting more random noise to the data, with the generative process learning to progressively remove noise to obtain realistic samples by mimicking the reverse process. 

% \textbf{Property 1.} Tractable marginal of the forward process.
% \begin{equation}
% \begin{aligned}
% q(\x_t | \x_0) &= \int q(\x_{1:t}|\x_0) \,d\x_{1:{(t-1)}}\\
% &= \N(\x_t; \sqrt{\tilde{\alpha}_t} \x_0, (1-\tilde{\alpha}_t) \I) \nonumber 
% \end{aligned}
% \end{equation}
% This property is proved in the Appendix of~\cite{ho2020denoising} and provides convenient closed-form evaluation of $\x_t$ knowing $\x_0$. That is, $\x_t = \sqrt{\tilde{\alpha}_t} \x_0 + \sqrt{1-\tilde{\alpha}_t} \bm{\epsilon}$, where $\bm{\epsilon} \sim \N(0, \I)$.
% Additionally, $\beta_T$ should be set such that $(1-\tilde{\alpha}_t)$ approaches 1 and $ \sqrt{\tilde{\alpha}_t}$ approaches 0 as to match $q(\x_T|x_0)$ with the Gaussian prior $p(\x_T)$.

% \textbf{Property 2.} Tractable posterior of the forward process.
% We first note the Bayes' rule that connects the posterior with the forward process,
% \[
% q(\x_{t-1} | \x_t, \x_0) = \frac{q(\x_t | \x_{t-1}, \x_0)q(\x_{t-1}|\x_0)}{q(\x_t |\x_0)} 
% \]
% and given the previous property, the posterior is also Gaussian, given by
% \begin{equation}\label{eq:posterior}
% \begin{aligned}
% q(\x_{t-1} | \x_t, \x_0) &= 
% \N(\x_{t-1};\frac{\sqrt{\tilde{\alpha}_{t-1}} \beta_t}{1-\tilde{\alpha}_t} \x_0 +\\ &\frac{\sqrt{\alpha_t} (1-\tilde{\alpha}_{t-1})}{1-\tilde{\alpha}_t} \x_t,
% \frac{(1-\tilde{\alpha}_{t-1})}{1-\tilde{\alpha}_t}\beta_t \I)  \end{aligned}
% \end{equation}

\myparagraph{Training objective.}
To learn the marginal likelihood $p_{\theta}(\x)$, we maximize a variational lower bound of log data likelihood that involves all of $\x_0, ...,\x_T$:\footnoteref{details}
\begin{equation}
    \E_{q(\x_0)}[\log p_\theta(\x_0)] \geq \E_{q(\x_{0:T})}\Big[\log \frac{p_\theta(\x_{0:T})}{q(\x_{1:T}|\x_0)}\Big].
\end{equation}
In the above objective, the forward process $q(\x_t | \x_{t-1})$ is fixed and $p(\x_T)$ is defined as a Gaussian prior, so they do not affect the learning of $\theta$. Therefore, the final objective can be reduced to maximum likelihood given the complete data likelihood with joint posterior $q(\x_{1:T}|\x_0)$:
\begin{equation}\label{eq:final-obj}
\max_{\theta} \E_{\x_0 \sim q(\x_0), \x_{1:T} \sim q(\x_{1:T}|\x_0)}\left[\sum_{t=1}^{T}\log p_\theta(\x_{t-1}|\x_t)\right].
\end{equation}

Joint posterior $q(\x_{1:T}|\x_0)$ can be factorized into $\prod_{t=1}^T{q(\x_{t-1} | \x_t, \x_0)}$. Each factored ground-truth posterior is denoted as $q(\x_{t-1} | \x_t, \x_0)$ and is analytically tractable. It can be shown that it is also parameterized by Gaussian distributions:
\begin{equation}\label{eq:posterior}
\begin{aligned}
& q(\x_{t-1} | \x_t, \x_0) = \\
&\N\left(\frac{\sqrt{\tilde{\alpha}_{t-1}} \beta_t}{1-\tilde{\alpha}_t} \x_0 + \frac{\sqrt{\alpha_t} (1-\tilde{\alpha}_{t-1})}{1-\tilde{\alpha}_t} \x_t,
\frac{(1-\tilde{\alpha}_{t-1})}{1-\tilde{\alpha}_t}\beta_t \I\right).  
\end{aligned}
\end{equation}
where $\alpha_t = 1 - \beta_t$ and $\tilde{\alpha}_t = \prod_{s=1}^{t} \alpha_s$.\footnoteref{details} This property allows each timestep to learn independently, \ie, each $p_\theta(\x_{t-1}|\x_t)$ only needs to match $q(\x_{t-1} | \x_t, \x_0)$.

Since both $p_\theta(\x_{t-1}|\x_t)$ and $q(\x_{t-1} | \x_t, \x_0)$ are Gaussian, we can reparameterize the model to output noise and the final loss can be reduced to an $\mathcal{L}_2$ loss between the model output $\bm{\epsilon}_\theta(\x_t, t)$ and noise $\bm{\epsilon}$:\footnoteref{details}
\begin{equation}\label{eq:l2-eps}
    \norm{\bm{\epsilon} - \bm{\epsilon}_\theta(\x_t, t)}^2,\;\; \bm{\epsilon} \sim \N(0, \I),
\end{equation}
Intuitively, the model seeks to predict the noise vector necessary to decorrupt the 3D shape.

Point clouds can then be generated by progressively sampling from $p_\theta(\x_{t-1}|\x_t)$ as $t=T,...,1$ using the following equation:
\begin{equation}\label{eq:generate}
    \x_{t-1} = \frac{1}{\sqrt{\alpha_t}}\left(\x_t - \frac{1 - \alpha_t}{\sqrt{1-\tilde{\alpha}_t}} \bm{\epsilon}_\theta(\x_t, t)\right) + \sqrt{\beta_t} \mathbf{z},
\end{equation}
where $\mathbf{z} \sim \N(0, \I)$, corresponding to the gradual denoising of a shape from noise.\footnoteref{details}

\subsection{Shape Completion}

Our objective can be simply modified to learn a conditional generative model given partial shapes, which we introduce in this section.

Denote a point cloud sample as $\x_0 = (\z_0, \tilde{\x}_0)$, where $\z_0 \in \mathbb{R}^{M\times 3}$ is the fixed partial shape, and any intermediate shapes as free points $\x_t = (\z_0, \tilde{\x}_t)$. We can then define a conditional forward process, where the partial shape is fixed at $\z_0$ for all time. Our conditional forward and generative processes, as well as each transition probability, can then be parametrized as 
\begin{equation}\label{eq:transition-partial}
\begin{aligned}
q(\tilde{\x}_t | \tilde{\x}_{t-1}, \z_0) &:= \N(\sqrt{1-\beta_t} \tilde{\x}_{t-1}, \beta_t \I),    \\
p_\theta( \tilde{\x}_{t-1}|\tilde{\x}_t, \z_0) &:= \N(\mu_\theta(\x_t, \z_0, t), \sigma_t^2  \I).
\end{aligned}
\end{equation}
Note that the above equations now give the forward/generative transition probabilities for the free points $\tilde{\x}_t$, while $\z_0$ stays unchanged for all timesteps. Intuitively, this process is the same as unconditional generation, while we hold the partial shape $\z_0$ fixed and diffuse only the missing parts.

The modified training objective also maximizes the likelihood conditioned on partial shapes $\z_0$:
\begin{equation}\label{eq:final-obj-partial}
\E_{(\tilde{\x}_0, \z_0) \sim q(x_0), \x_{1:T} \sim q(\x_{1:T}|\tilde{\x}_0,\z_0)}\left[\sum_{t=1}^{T}\log p_\theta(\tilde{\x}_{t-1}|\tilde{\x}_t, \z_0)\right],
\end{equation}
where each posterior $q(\tilde{\x}_{t-1} | \tilde{\x}_t, \tilde{\x}_0, \z_0)$ is known and its derivation is similar to the unconditional generative model. Using the same reasoning as before, we can arrive at a similar $\mathcal{L}_2$ loss:
\begin{equation}\label{eq:l2-eps-partial}
\mathcal{L}_t = \norm{\bm{\epsilon} - \bm{\epsilon}_\theta(\tilde{\x}_t, \z_0, t)}^2,
\end{equation}
where $\bm{\epsilon} \sim \N(0, \I)$. Additionally, since the partial shape is always fixed during both forward and generative processes, we can mask away the subset of model output that affects $\z_0$ and minimize $\mathcal{L}_2$ distance between $\tilde{\bm{\epsilon}}(\tilde{\x}_t, \z_0, t)$ and random noise, which only affects $\tilde{\x}_t$. In practice, we input $\z_0$ and $\x_t$ into the model and obtain $\x_{t-1}$, where only the subset $\tilde{\x}_{t-1}$ is used for $\mathcal{L}_2$ loss. In shape completion, $\tilde{\x}_{t-1}$ is concatenated with $\z_0$ to be the input into the model again. This allows the exact same training architecture to do both generation and shape completion by simply changing the training objective.

\begin{figure*}[t]
\centering

\begin{tabular}{ccccc|cc|cc}
%\noalign{\vskip 0.1cm}
\begin{subfigure}[b]{0.09\textwidth}
\centering
\includegraphics[width=\linewidth]{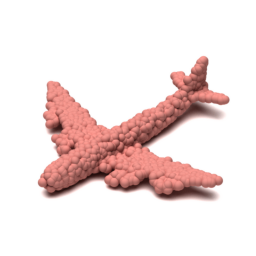}
\end{subfigure}
            &
\begin{subfigure}[b]{0.09\textwidth}
\centering
\includegraphics[width=\linewidth]{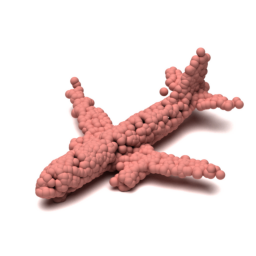}
\end{subfigure}
            &
\begin{subfigure}[b]{0.09\textwidth}
\centering
\includegraphics[width=\linewidth]{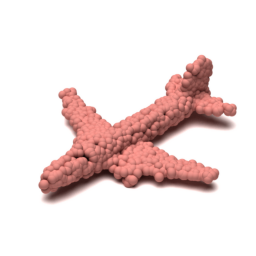}
\end{subfigure}
            &
\begin{subfigure}[b]{0.09\textwidth}
\centering
\includegraphics[width=\linewidth]{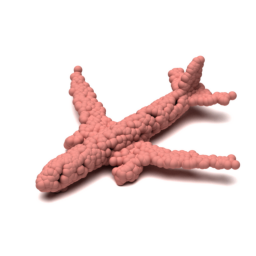}
\end{subfigure}
&
\begin{subfigure}[b]{0.09\textwidth}
\centering
\includegraphics[width=\linewidth]{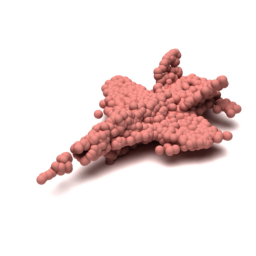}
\end{subfigure}
&
\begin{subfigure}[b]{0.09\textwidth}
\centering
\includegraphics[width=\linewidth]{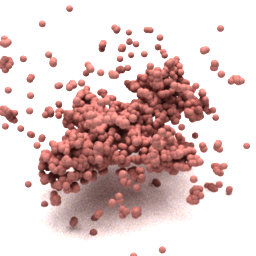}
\end{subfigure}
&
\begin{subfigure}[b]{0.09\textwidth}
\centering
\includegraphics[width=\linewidth]{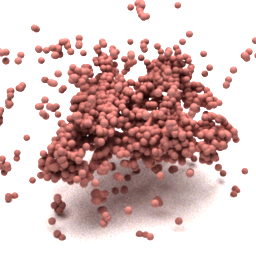}
\end{subfigure}
&
\begin{subfigure}[b]{0.09\textwidth}
\centering
\includegraphics[width=\linewidth]{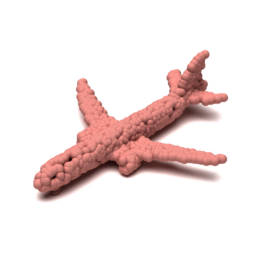}
\end{subfigure}
&
\begin{subfigure}[b]{0.09\textwidth}
\centering
\includegraphics[width=\linewidth]{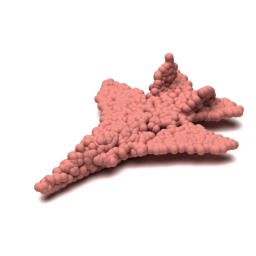}
\end{subfigure}
            \\
\begin{subfigure}[b]{0.09\textwidth}
\centering
\includegraphics[width=\linewidth]{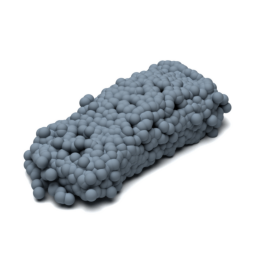}
\end{subfigure}
            &
\begin{subfigure}[b]{0.09\textwidth}
\centering
\includegraphics[width=\linewidth]{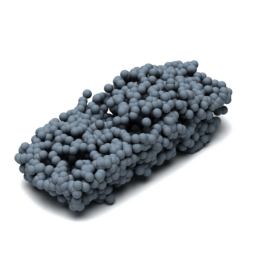}
\end{subfigure}
            &
\begin{subfigure}[b]{0.09\textwidth}
\centering
\includegraphics[width=\linewidth]{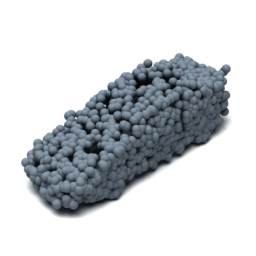}
\end{subfigure}
            &
\begin{subfigure}[b]{0.09\textwidth}
\centering
\includegraphics[width=\linewidth]{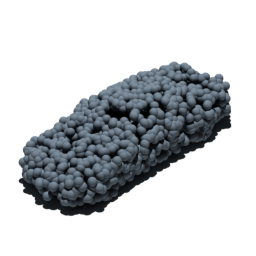}
\end{subfigure}
&
\begin{subfigure}[b]{0.09\textwidth}
\centering
\includegraphics[width=\linewidth]{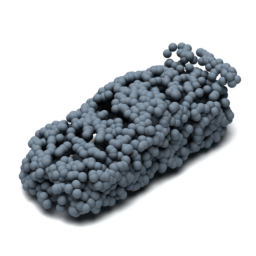}
\end{subfigure}
&
\begin{subfigure}[b]{0.09\textwidth}
\centering
\includegraphics[width=\linewidth]{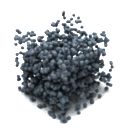}
\end{subfigure}
&
\begin{subfigure}[b]{0.09\textwidth}
\centering
\includegraphics[width=\linewidth]{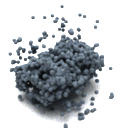}
\end{subfigure}
&
\begin{subfigure}[b]{0.09\textwidth}
\centering
\includegraphics[width=\linewidth]{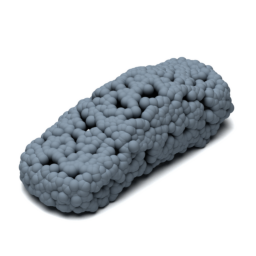}
\end{subfigure}
&
\begin{subfigure}[b]{0.09\textwidth}
\centering
\includegraphics[width=\linewidth]{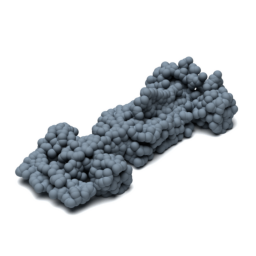}
\end{subfigure}
              \\
\begin{subfigure}[b]{0.09\textwidth}
\centering
\includegraphics[width=\linewidth]{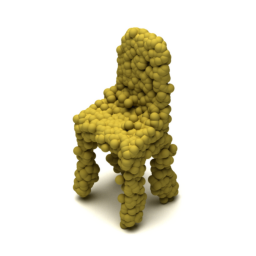}
\end{subfigure}
            &
\begin{subfigure}[b]{0.09\textwidth}
\centering
\includegraphics[width=\linewidth]{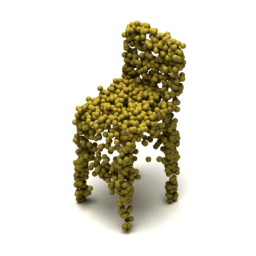}
\end{subfigure}
            &
\begin{subfigure}[b]{0.09\textwidth}
\centering
\includegraphics[width=\linewidth]{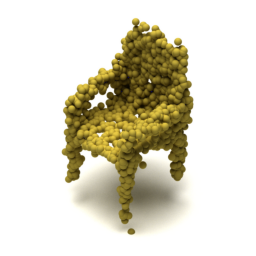}
\end{subfigure}
            &
\begin{subfigure}[b]{0.09\textwidth}
\centering
\includegraphics[width=\linewidth]{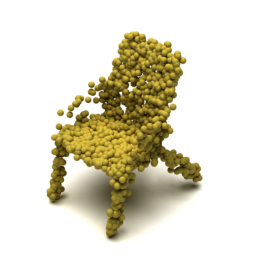}
\end{subfigure}
&
\begin{subfigure}[b]{0.09\textwidth}
\centering
\includegraphics[width=\linewidth]{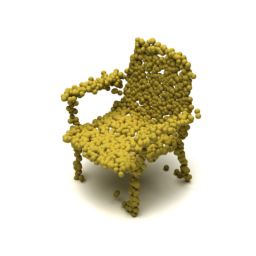}
\end{subfigure}
&
\begin{subfigure}[b]{0.09\textwidth}
\centering
\includegraphics[width=\linewidth]{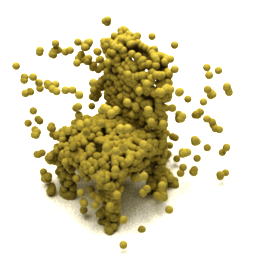}
\end{subfigure}
&
\begin{subfigure}[b]{0.09\textwidth}
\centering
\includegraphics[width=\linewidth]{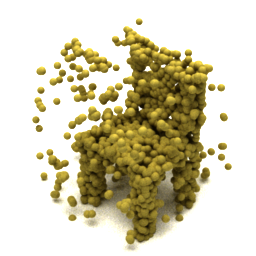}
\end{subfigure}
&
\begin{subfigure}[b]{0.09\textwidth}
\centering
\includegraphics[width=\linewidth]{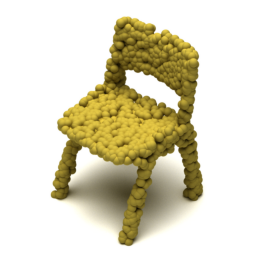}
\end{subfigure}
&
\begin{subfigure}[b]{0.09\textwidth}
\centering
\includegraphics[width=\linewidth]{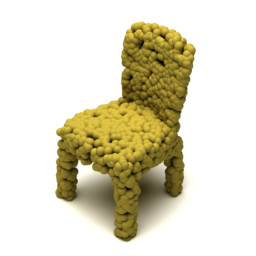}
\end{subfigure}
            \\
l-GAN &PointFlow & SoftFlow & DPF-Net & Shape-GF &\multicolumn{2}{>{\centering}c}{Vox-Diff} &\multicolumn{2}{>{\centering}c}{\model (ours)}
\end{tabular}
\caption{Results on unconditional shape generation with 2,048 points. The l-GAN results are from the EMD variant.}
\label{fig:gen}

\end{figure*}%

\section{Experiments}

We demonstrate here that our model outperforms previous point generative models in \sect{sect:shape_gen}, is capable of completing partial shapes sampled from single views in \sect{sect:shape_complete}, and can generate diverse shapes given partial shape constraints in \sect{sect:multi_modal}. Architecture and hyper-parameter details are provided in Appendix.

%\jw{Let's also cite ShapeNet}
\subsection{Shape Generation}
\label{sect:shape_gen}

\begin{table}[t]
\centering
\small
\setlength{\tabcolsep}{4pt}
\begin{tabular}{lcccccc}
\toprule
            & \multicolumn{2}{c}{Airplane}             & \multicolumn{2}{c}{Chair}                 & \multicolumn{2}{c}{Car}         \\ 
\cmidrule(lr){2-3}\cmidrule(lr){4-5}\cmidrule(lr){6-7}
            % & \multicolumn{2}{c|}{1-NN, \%}             & \multicolumn{2}{c|}{1-NN, \%}             & \multicolumn{2}{c}{1-NN, \%}    \\ \cline{2-7} 
            & CD             & \multicolumn{1}{c}{EMD} & CD             & \multicolumn{1}{c}{EMD} & CD             & EMD            \\ \midrule
r-GAN~\cite{achlioptas2018learning}       & 98.40          & 96.79                    & 83.69          & 99.70                    & 94.46          & 99.01          \\
l-GAN (CD)~\cite{achlioptas2018learning}  & 87.30          & 93.95                    & 68.58          & 83.84                    & 66.49          & 88.78          \\
l-GAN (EMD)~\cite{achlioptas2018learning} & 89.49          & 76.91                    & 71.90          & 64.65                    & 71.16          & 66.19          \\
% SetVAE~\cite{Kim_2021_CVPR}   &    76.17     &    68.64        &   57.85   &    59.96      &    58.38       &    56.81     \\
% HyperCloud~\cite{spurek20a}   & 93.70  &   90.98 &    68.20  &      68.80     &    86.93   &  78.38   \\
PointFLow~\cite{yang2019pointflow}   & 75.68          & 70.74                    & 62.84          & 60.57                    & 58.10          & 56.25          \\
SoftFlow~\cite{kim2020softflow}    & 76.05          & 65.80                    & 59.21          & 60.05                    & 64.77          & 60.09          \\
DPF-Net~\cite{klokov2020discrete}     & 75.18          & 65.55                    & 62.00          & 58.53                    & 62.35          & 54.48          \\
Shape-GF~\cite{cai2020learning}    & 80.00          & 76.17                    & 68.96          & 65.48                    & 63.20          & 56.53          \\
\model (ours)   & \textbf{73.82} & \textbf{64.81}           & \textbf{56.26} & \textbf{53.32}           & \textbf{54.55} & \textbf{53.83}\\
\bottomrule
\end{tabular}
\caption{Generation results on Airplane, Chair, Car compared with baselines using 1-NN as the metric. Both CD and EMD as the distance measure are calculated. Lower scores indicate better quality and diversity.}
\label{tab:gen}
\end{table}
\paragraph{Data.} We choose ShapeNet~\cite{chang2015shapenet} Airplane, Chair, and Car to be our main datasets for generation, following most previous works~\cite{yang2019pointflow, klokov2020discrete, cai2020learning, kim2020softflow}. We use the provided datasets in~\cite{yang2019pointflow}, which contain 15,000 sampled points for each shape. We sample 2,048 points for training and testing, respectively, and process our data following procedures provided in PointFlow~\cite{yang2019pointflow}. 

\myparagraph{Evaluation metrics.} Previous works such as~\cite{yang2019pointflow, klokov2020discrete, cai2020learning, kim2020softflow} have used Chamfer Distance (CD) and Earth Mover’s Distance (EMD) as their distance metrics in calculating Jensen-Shannon Divergence (JSD), Coverage (COV), Minimum Matching Distance (MMD), and 1-Nearest Neighbor (1-NN), which are four main metrics to measure generative quality. However, as discussed by~\cite{yang2019pointflow}, JSD, COV, and MMD each has limitations and does not necessarily indicate better quality. Some generation results achieve even better scores than ground-truth datasets. 
1-NN is robust and correlates with generation quality, as supported by~\cite{yang2019pointflow}, which also proposes 1-NN as the better metric. Therefore, we use 1-NN directly for evaluating generation quality and we provide comparison of remaining metrics in Appendix. As we also discover that EMD score can vary widely depending on its implementation, We evaluate all baselines using our implementation of the metrics.

\myparagraph{Baselines and results.} We quantitatively compare our results with r-GAN~\cite{achlioptas2018learning}, l-GAN~\cite{achlioptas2018learning}, PointFlow~\cite{yang2019pointflow}, DPF-Net~\cite{klokov2020discrete}, SoftFlow~\cite{kim2020softflow}, and Shape-GF~\cite{cai2020learning} on generating 2048 points. In evaluating the baselines, we follow the same data processing and evaluation procedure as PointFlow, and follow the provided baseline implementations to evaluate their models. Our comparisons are shown in Table~\ref{tab:gen}. Our model noticeably achieves the best generation quality. 

We also investigated pure voxel and point representations for shape generation. We noticed that simply extending diffusion models to pure point representation using conventional permutation-invariant architectures such as PointNet++~\cite{qi2017pointnet++} fails to generate any visible shapes. Extending diffusion models to pure voxel representation generates noisy results due to the binary nature of voxels which is different from our Gaussian assumption. We visually compare with baselines including a voxel diffusion model (Vox-Diff) in Figure~\ref{fig:gen}. For Vox-Diff, 2048 points are sampled from voxel surfaces. We provide additional quantitative comparison in Appendix.

\begin{figure*}[t]
\centering

\begin{tabular}{ccc|ccc|c}
%\noalign{\vskip 0.1cm}
\begin{subfigure}[b]{0.10\textwidth}
\centering
\includegraphics[width=\linewidth]{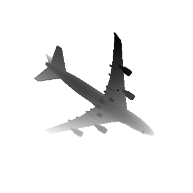}
\end{subfigure}
            &
\begin{subfigure}[b]{0.10\textwidth}
\centering
\includegraphics[width=\linewidth]{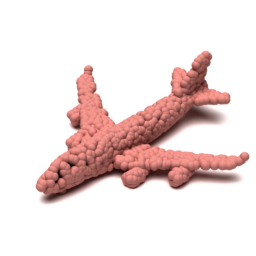}
\end{subfigure}
            &
\begin{subfigure}[b]{0.10\textwidth}
\centering
\includegraphics[width=\linewidth]{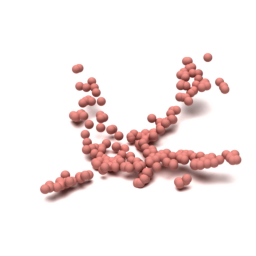}
\end{subfigure}
&
\begin{subfigure}[b]{0.10\textwidth}
\centering
\includegraphics[width=\linewidth]{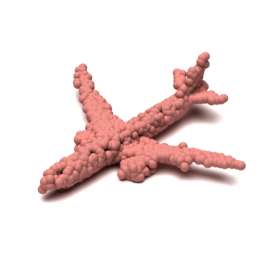}
\end{subfigure}
&
\begin{subfigure}[b]{0.10\textwidth}
\centering
\includegraphics[width=\linewidth]{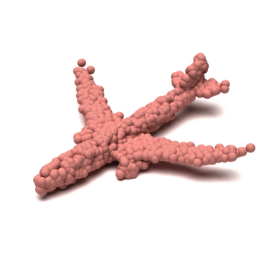}
\end{subfigure}
&
\begin{subfigure}[b]{0.10\textwidth}
\centering
\includegraphics[width=\linewidth]{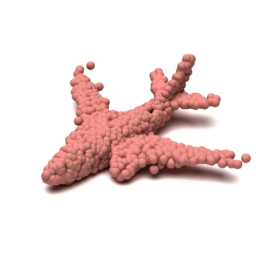}
\end{subfigure}

% &
% \begin{subfigure}[b]{0.085\textwidth}
% \centering
% \includegraphics[width=\linewidth]{figures/completion/msn_sc_airplane.png}
% \end{subfigure}
% &
% \begin{subfigure}[b]{0.085\textwidth}
% \centering
% \includegraphics[width=\linewidth]{figures/completion/pcn_sc_airplane.png}
% \end{subfigure}
&
\begin{subfigure}[b]{0.10\textwidth}
\centering
\includegraphics[width=\linewidth]{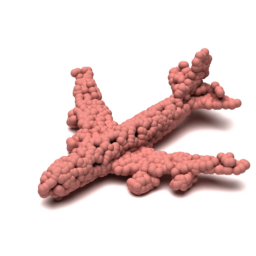}
\end{subfigure}
            \\

\begin{subfigure}[b]{0.10\textwidth}
\centering
\includegraphics[width=\linewidth]{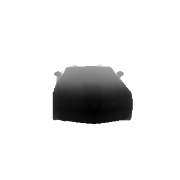}
\end{subfigure}
            &
\begin{subfigure}[b]{0.10\textwidth}
\centering
\includegraphics[width=\linewidth]{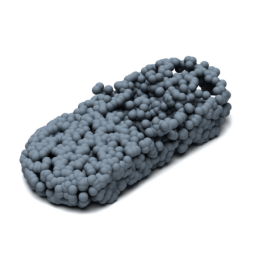}
\end{subfigure}
            &
\begin{subfigure}[b]{0.10\textwidth}
\centering
\includegraphics[width=\linewidth]{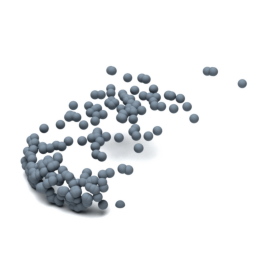}
\end{subfigure}
&
\begin{subfigure}[b]{0.10\textwidth}
\centering
\includegraphics[width=\linewidth]{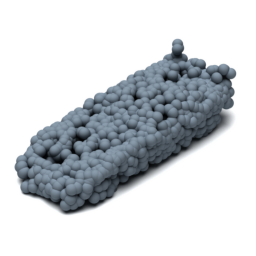}
\end{subfigure}
&
\begin{subfigure}[b]{0.10\textwidth}
\centering
\includegraphics[width=\linewidth]{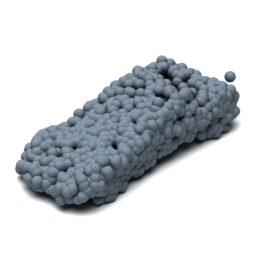}
\end{subfigure}
&
\begin{subfigure}[b]{0.10\textwidth}
\centering
\includegraphics[width=\linewidth]{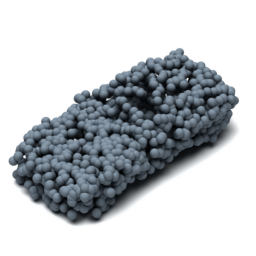}
\end{subfigure}
% &
% \begin{subfigure}[b]{0.085\textwidth}
% \centering
% \includegraphics[width=\linewidth]{figures/completion/msn_sc_car.png}
% \end{subfigure}
% &
% \begin{subfigure}[b]{0.085\textwidth}
% \centering
% \includegraphics[width=\linewidth]{figures/completion/pcn_sc_car.png}
% \end{subfigure}

&
\begin{subfigure}[b]{0.10\textwidth}
\centering
\includegraphics[width=\linewidth]{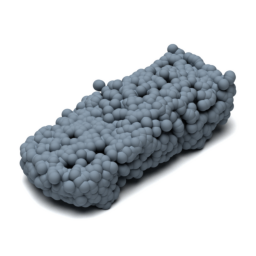}
\end{subfigure}

            \\
\begin{subfigure}[b]{0.10\textwidth}
\centering
\includegraphics[width=\linewidth]{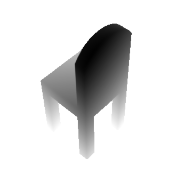}
\end{subfigure}
            &
\begin{subfigure}[b]{0.10\textwidth}
\centering
\includegraphics[width=\linewidth]{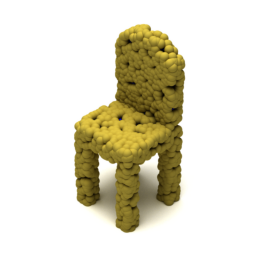}
\end{subfigure}
            &
\begin{subfigure}[b]{0.10\textwidth}
\centering
\includegraphics[width=\linewidth]{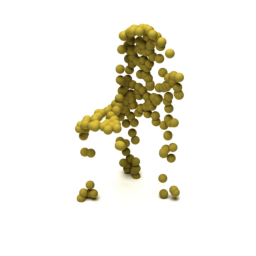}
\end{subfigure}
&
\begin{subfigure}[b]{0.10\textwidth}
\centering
\includegraphics[width=\linewidth]{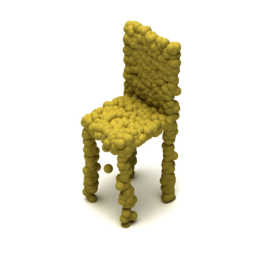}
\end{subfigure}
&
\begin{subfigure}[b]{0.10\textwidth}
\centering
\includegraphics[width=\linewidth]{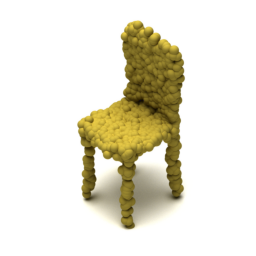}
\end{subfigure}
&
\begin{subfigure}[b]{0.10\textwidth}
\centering
\includegraphics[width=\linewidth]{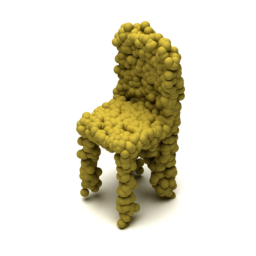}
\end{subfigure}

% &
% \begin{subfigure}[b]{0.085\textwidth}
% \centering
% \includegraphics[width=\linewidth]{figures/completion/msn_sc_chair.png}
% \end{subfigure}
% &
% \begin{subfigure}[b]{0.085\textwidth}
% \centering
% \includegraphics[width=\linewidth]{figures/completion/pcn_sc_chair.png}
% \end{subfigure}
&
\begin{subfigure}[b]{0.10\textwidth}
\centering
\includegraphics[width=\linewidth]{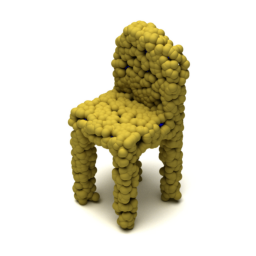}
\end{subfigure}
            \\

Depths & Ground-truths & Partial & PointFlow & SoftFlow & DPF-Net & \model (ours)
    
\end{tabular}
\caption{Our shape completion visualization (right) compared to baseline models (left). From left to right: depth images, ground-truth shapes, partial shapes sampled from depth images, completion from baselines, and our results.}
\label{fig:sc}

\end{figure*}%

\subsection{Shape Completion}
\label{sect:shape_complete}

In various graphics applications users usually do not have access to all viewpoints of an object. A shape often needs to be completed knowing a partial shape from a single depth map. Therefore, the ability to complete partial shapes become practically useful. In this section, we use the same model architecture (see Appendix) from Section~\ref{sect:shape_gen} and test our shape completion models. 

\myparagraph{Data.} For shape completion, we use the benchmark provided by GenRe~\cite{zhang2018learning}, which contains renderings of each shape in ShapeNet from 20 random views. We sample 200 points as our partial point clouds obtained from the provided depth images, and we evaluate shape completion on all 20 partial shapes per ground-truth sample. 

\myparagraph{Metrics.} For shape completion, as the ground-truth data are involved, Chamfer Distance and Earth Mover's Distance suffice to evaluate the reconstruction results.

\myparagraph{Baselines.} Since our approach is probabilistic, we selected major distribution-fitting models such as PointFlow~\cite{yang2019pointflow}, DPF-Net~\cite{klokov2020discrete}, and  SoftFlow~\cite{kim2020softflow} for comparison. We directly evaluate pre-trained models, if provided, otherwise we re-train them using baselines' provided implementation on our benchmark. We also compared with Shape-GF as its encoder can similarly receive an arbitrary number of points. However, it is experimentally found that the model is sensitive to the input partial shapes and completion is not realistic after Langevin sampling. Therefore, we leave them out of the comparison. %\jw{why not just include the numbers? It's fine if they are bad.} 

%\jw{I think it's hard to argue against PCN or MSN. I suggest we remove them as baselines and also remove Table 3. Essentially they are not developed to model the distribution of shapes..}

\begin{table}[t]
\centering
\begin{tabular}{clcc}
\toprule
Category  & Model &   CD   &   EMD  \\
\midrule
\multirow{4}{*}{Airplane} 
                  &   SoftFlow~\cite{kim2020softflow}    &  0.4042   & 1.198  \\  
                  &   PointFlow~\cite{yang2019pointflow}   &  \textbf{0.4030}  & 1.180  \\
                  &   DPF-Net~\cite{klokov2020discrete} &  0.5279 &  1.105  \\  
                  &   \model (ours) &  0.4415  &  \textbf{1.030}  \\
\midrule
\multirow{4}{*}{Chair}  
                  &   SoftFlow~\cite{kim2020softflow}   &   2.786  & 3.295   \\ 
                  &   PointFlow~\cite{yang2019pointflow}   &  \textbf{2.707}  &  3.649   \\ 
                  &   DPF-Net~\cite{klokov2020discrete}     & 2.763   &  3.320   \\  
                  &   \model (ours)   &  3.211  &  \textbf{2.939}\\
\midrule
\multirow{4}{*}{Car} 
                  &   SoftFlow~\cite{kim2020softflow}   &   1.850   &   2.789  \\ 
                  &   PointFlow~\cite{yang2019pointflow}   &  1.803   &   2.851 \\  
                  &   DPF-Net~\cite{klokov2020discrete}   & \textbf{1.396}  &  2.318  \\ 
                  &   \model (ours) &    1.774 & \textbf{2.146}  \\
\bottomrule
\end{tabular}
\caption{Quantitative comparison against baselines. CD is multiplied by $10^3$ and EMD is multiplied by  $10^2$.}
\label{tab:sc-1}
\end{table}

\begin{figure}[t]
\centering
\includegraphics[width=\linewidth]{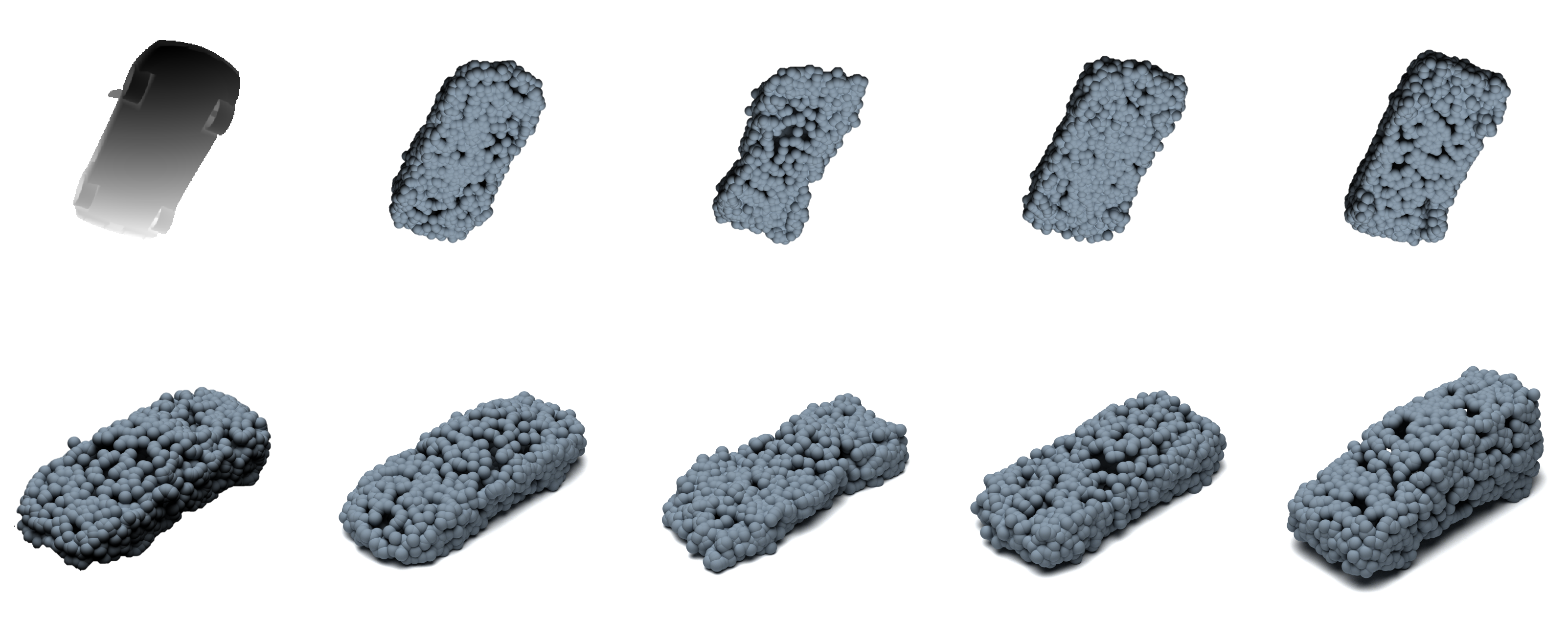}
\setlength{\tabcolsep}{6pt}
\begin{tabular}{ccccc}
GT & PointFlow & SoftFlow & DPF-Net & \model (ours) \\
\cmidrule(r){1-5}
CD  & 8.633 & 1.501 & 1.197 & 1.921 \\
EMD & 1.124 & 1.856 & 1.109 & 0.919 \\ %0.9192 \\
\end{tabular}
\caption{Typical case when CD is higher than baseline models. Column 1 shows input depth image and ground-truth point clouds. The next columns show completion from the input viewpoint (top) and from the canonical viewpoint (bottom). CD is multiplied by $10^3$ and EMD is multiplied by $10^2$ scores.}
\label{fig:sc-failure}
\end{figure}

\myparagraph{Results.} Quantitative results are presented in Table~\ref{tab:sc-1} and a visual comparison is shown in Figure~\ref{fig:sc}. From Table~\ref{tab:sc-1}, we observe that our model achieves best on EMD scores while worse on CD compared to some baselines. First, we note EMD is a better metric for measuring completion quality because by solving the linear assignment problem it forces model outputs to have the same density as the ground-truths~\cite{liu2020morphing} and it is known that CD is blind to visual inferiority~\cite{achlioptas2018learning}. Our better EMD score is more indicative of higher visual quality.
%From Table~\ref{tab:sc-2}, our model is not as quantitatively competitive as MSN and PCN in primarily because, (1) they directly use EMD and CD as their optimization objectives, and (2) they learn to output major modes from the dataset as mean shapes, which are naturally close to ground-truths than more nuanced completions. Figure~\ref{fig:sc-pcn-msn} shows that on two shapes with small but important differences, our model produces more detailed completion while the baselines' outputs are very similar. Notice that in the figure PCN and MSN can achieve lower scores than ours even though visually their completion is not as faithful.

\begin{figure*}[t]
\centering
\begin{tabular}{p{0.05\textwidth}p{0.05\textwidth}p{0.05\textwidth}p{0.05\textwidth}p{0.05\textwidth}p{0.05\textwidth}p{0.05\textwidth}p{0.05\textwidth}p{0.05\textwidth}p{0.05\textwidth}p{0.05\textwidth}p{0.05\textwidth}}
\multirow{5}{*}{\begin{subfigure}[t]{0.05\textwidth}\vspace*{1cm}
\centering
\includegraphics[width=\linewidth]{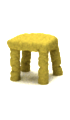}
\end{subfigure}} & \begin{subfigure}[t]{0.05\textwidth}
\centering
\includegraphics[width=\linewidth]{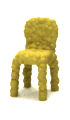}
\end{subfigure} & \begin{subfigure}[t]{0.05\textwidth}
\centering
\includegraphics[width=\linewidth]{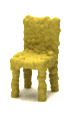}
\end{subfigure} & \begin{subfigure}[t]{0.05\textwidth}
\centering
\includegraphics[width=\linewidth]{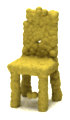}
\end{subfigure} & \multirow{5}{*}{\begin{subfigure}[t]{0.05\textwidth}\vspace*{1.2cm}
\centering
\includegraphics[width=\linewidth]{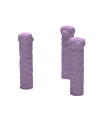}
\end{subfigure}} & \begin{subfigure}[t]{0.05\textwidth}
\centering
\includegraphics[width=\linewidth]{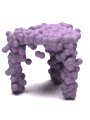}
\end{subfigure} & \begin{subfigure}[t]{0.05\textwidth}
\centering
\includegraphics[width=\linewidth]{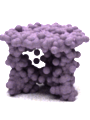}
\end{subfigure} &  \begin{subfigure}[t]{0.05\textwidth}
\centering
\includegraphics[width=\linewidth]{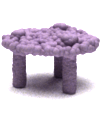}
\end{subfigure}& \multirow{5}{*}{\begin{subfigure}[t]{0.05\textwidth}\vspace*{1cm}
\centering
\includegraphics[width=\linewidth]{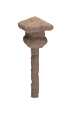}
\end{subfigure}} & \begin{subfigure}[t]{0.05\textwidth}
\centering
\includegraphics[width=\linewidth]{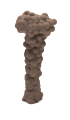}
\end{subfigure}  & \begin{subfigure}[t]{0.05\textwidth}
\centering
\includegraphics[width=\linewidth]{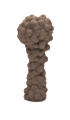}
\end{subfigure} & \begin{subfigure}[t]{0.05\textwidth}
\centering
\includegraphics[width=\linewidth]{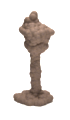}
\end{subfigure} \\
                  &\begin{subfigure}[t]{0.05\textwidth}
\centering
\includegraphics[width=\linewidth]{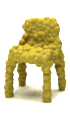}
\end{subfigure}  & \begin{subfigure}[t]{0.05\textwidth}
\centering
\includegraphics[width=\linewidth]{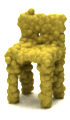}
\end{subfigure} & \begin{subfigure}[t]{0.05\textwidth}
\centering
\includegraphics[width=\linewidth]{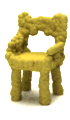}
\end{subfigure} &                   & \begin{subfigure}[t]{0.05\textwidth}
\centering
\includegraphics[width=\linewidth]{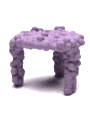}
\end{subfigure} & \begin{subfigure}[t]{0.05\textwidth}
\centering
\includegraphics[width=\linewidth]{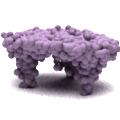}
\end{subfigure} & \begin{subfigure}[t]{0.05\textwidth}
\centering
\includegraphics[width=\linewidth]{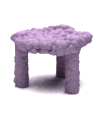}
\end{subfigure} &                   & \begin{subfigure}[t]{0.05\textwidth}
\centering
\includegraphics[width=\linewidth]{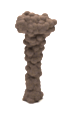}
\end{subfigure} & \begin{subfigure}[t]{0.05\textwidth}
\centering
\includegraphics[width=\linewidth]{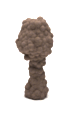}
\end{subfigure} & \begin{subfigure}[t]{0.05\textwidth}
\centering
\includegraphics[width=\linewidth]{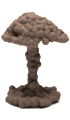}
\end{subfigure} \\
                  & \begin{subfigure}[t]{0.05\textwidth}
\centering
\includegraphics[width=\linewidth]{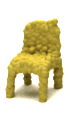}
\end{subfigure} & \begin{subfigure}[t]{0.05\textwidth}
\centering
\includegraphics[width=\linewidth]{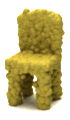}
\end{subfigure} & \begin{subfigure}[t]{0.05\textwidth}
\centering
\includegraphics[width=\linewidth]{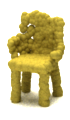}
\end{subfigure} &                   & \begin{subfigure}[t]{0.05\textwidth}
\centering
\includegraphics[width=\linewidth]{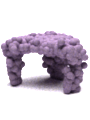}
\end{subfigure} & \begin{subfigure}[t]{0.05\textwidth}
\centering
\includegraphics[width=\linewidth]{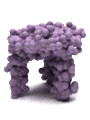}
\end{subfigure}  & \begin{subfigure}[t]{0.05\textwidth}
\centering
\includegraphics[width=\linewidth]{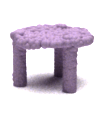}
\end{subfigure} &                   &  \begin{subfigure}[t]{0.05\textwidth}
\centering
\includegraphics[width=\linewidth]{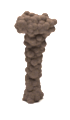}
\end{subfigure}& \begin{subfigure}[t]{0.05\textwidth}
\centering
\includegraphics[width=\linewidth]{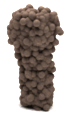}
\end{subfigure} & \begin{subfigure}[t]{0.05\textwidth}
\centering
\includegraphics[width=\linewidth]{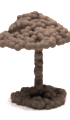}
\end{subfigure} \\
                  &\begin{subfigure}[t]{0.05\textwidth}
\centering
\includegraphics[width=\linewidth]{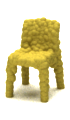}
\end{subfigure}  & \begin{subfigure}[t]{0.05\textwidth}
\centering
\includegraphics[width=\linewidth]{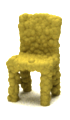}
\end{subfigure} & \begin{subfigure}[t]{0.05\textwidth}
\centering
\includegraphics[width=\linewidth]{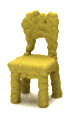}
\end{subfigure} &                   & \begin{subfigure}[t]{0.05\textwidth}
\centering
\includegraphics[width=\linewidth]{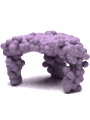}
\end{subfigure} & \begin{subfigure}[t]{0.05\textwidth}
\centering
\includegraphics[width=\linewidth]{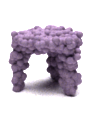}
\end{subfigure} & \begin{subfigure}[t]{0.05\textwidth}
\centering
\includegraphics[width=\linewidth]{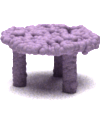}
\end{subfigure} &                   & \begin{subfigure}[t]{0.05\textwidth}
\centering
\includegraphics[width=\linewidth]{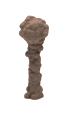}
\end{subfigure} &\begin{subfigure}[t]{0.05\textwidth}
\centering
\includegraphics[width=\linewidth]{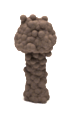}
\end{subfigure}  &  \begin{subfigure}[t]{0.05\textwidth}
\centering
\includegraphics[width=\linewidth]{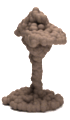}
\end{subfigure}\\
                  & \begin{subfigure}[t]{0.05\textwidth}
\centering
\includegraphics[width=\linewidth]{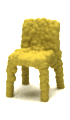}
\end{subfigure} & \begin{subfigure}[t]{0.05\textwidth}
\centering
\includegraphics[width=\linewidth]{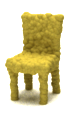}
\end{subfigure} & \begin{subfigure}[t]{0.05\textwidth}
\centering
\includegraphics[width=\linewidth]{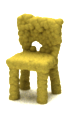}
\end{subfigure} &                   & \begin{subfigure}[t]{0.05\textwidth}
\centering
\includegraphics[width=\linewidth]{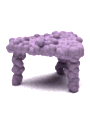}
\end{subfigure} & \begin{subfigure}[t]{0.05\textwidth}
\centering
\includegraphics[width=\linewidth]{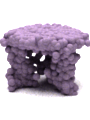}
\end{subfigure} & \begin{subfigure}[t]{0.05\textwidth}
\centering
\includegraphics[width=\linewidth]{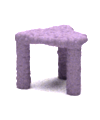}
\end{subfigure} &                   & \begin{subfigure}[t]{0.05\textwidth}
\centering
\includegraphics[width=\linewidth]{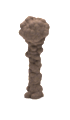}
\end{subfigure} & \begin{subfigure}[t]{0.05\textwidth}
\centering
\includegraphics[width=\linewidth]{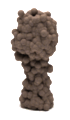}
\end{subfigure} & \begin{subfigure}[t]{0.05\textwidth}
\centering
\includegraphics[width=\linewidth]{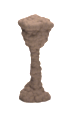}
\end{subfigure} \\
            Partial& KNN-latent  & cGAN & \model (Ours) & Partial &  KNN-latent & cGAN & \model (Ours) &   Partial &  KNN-latent  & cGAN & \model (Ours)
\end{tabular}
\caption{Multi-modal completion visualization on PartNet.  Each column presents five completion modes from a model.}
%\vspace{5pt}
\label{fig:mm_partnet}
\end{figure*}
Next, we investigate the reason our CD is inferior to our baselines. We discover that a typical case when our CD is higher is as shown in Figure~\ref{fig:sc-failure}, where from the input view the ground-truth shape is largely unknown. The baseline models tend to output a mean shape when encountered with such an unconventional angle. Naturally, mean shapes are more frequently closer to the ground-truths than other shapes, as exemplified by the figure. However, with each noise initialization, our model seeks a possible completion that matches well with the partial shape and may be further away from the ground-truth than the mean shape. In the case shown, our completion is a van instead of a sedan but is equally realistic.

Our model also enables controlled completion given multiple partial shapes, and we leave details to Appendix.

\begin{table}[t]
\centering\small
\setlength{\tabcolsep}{2pt}
\begin{tabular}{lcccccccc}
\toprule
                  & \multicolumn{4}{c}{TMD $\times 10^2$}                                  & \multicolumn{4}{c}{MMD $\times 10^3$}              \\
                  \cmidrule(lr){2-5}\cmidrule(lr){6-9}
                  & \multicolumn{1}{c}{Chair} & Table & Lamp & Avg.          & Chair & Table & Lamp & Avg.          \\
                  \midrule
KNN-latent~\cite{wu2020multimodal}       & 0.96                      & 1.37  & 1.95 & 1.43          & 1.42  & 1.42  & 1.88 & 1.57          \\
cGAN~\cite{wu2020multimodal}              & 1.75                      & 1.99  & 1.94 & 1.89          & 1.61  & 1.56  & 2.13 & 1.77          \\
\model (ours) & 1.91                      & 1.70  & 5.92 & \textbf{3.18} & 1.27  & 1.03  & 1.98 & \textbf{1.43}\\
\bottomrule
\end{tabular}
\caption{Quantitative comparison for multi-modal completion on PartNet. TMD (higher the better) measures completion diversity, and MMD (lower the better) measures completion quality. Chamfer Distance (CD) is used as the distance measure.}
\label{tab:mm-diversity}
\end{table}

\subsection{Multi-Modal Completion}
\label{sect:multi_modal}

Our baselines for shape completion adopt an encoder-decoder structure that takes in a partial shape and outputs a single completion. While some offer impressive results, their completion ability is deterministic, much different from humans who can often imagine different completion possibilities given single views. Our \model, however, adopts a probabilistic approach to shape completion, where each noise initialization can result in a different completion. %In this section, we demonstrate our model's multi-modal completion ability.

\myparagraph{Data.} We follow experiment setups from cGAN~\cite{wu2020multimodal} and train our model on Chair, Table, and Lamp from PartNet~\cite{Mo_2019_CVPR}. 1024 points are given and 2048 points are generated as completion. In addition, we show diversity of our completion on ShapeNet. Different from PartNet, ShapeNet's partial shapes are sampled from custom-rendered depth images.

\myparagraph{Metrics.} We follow cGAN~\cite{wu2020multimodal} and use Total Mutual Difference (TMD) to measure diversity and Minimal Matching Distance (MMD) to measure quality with Chamfer Distance (CD) as distance measure. Since our model is different from cGAN and only completes the 1024 free points, TMD is calculated only on the free points for our model and on a subsampled set of 1024 points for our baselines. MMD is calculated using the completed 2048 points and re-sampled 2048 ground-truth points.

\myparagraph{Results on PartNet.} Our model is compared with cGAN~\cite{wu2020multimodal} and KNN-latent~\cite{wu2020multimodal}. Results are shown in Table~\ref{tab:mm-diversity} and visual comparisons are shown in Figure~\ref{fig:mm_partnet}. Our model outperforms both baselines in terms of average diversity and quality.

\begin{figure*}[t]
\centering
\includegraphics[width=\linewidth]{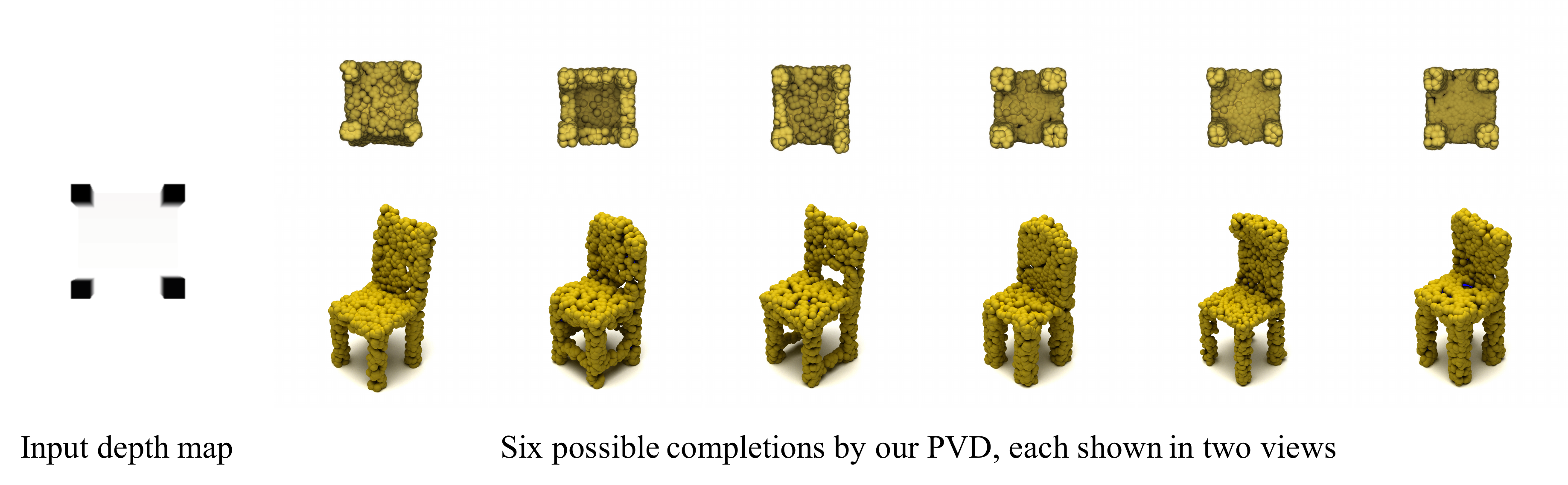}
\vspace{-20pt}
\setlength{\belowcaptionskip}{0pt}
\caption{Multi-modal completion results on ShapeNet. Left: ground-truth bottom view depth image of a chair. Right: six different possible shape completion results. Top: completion from the depth image viewpoint. Bottom: completion from the canonical viewpoint.}
%\vspace{5pt}
\label{fig:mm_shapenet}
\end{figure*}

\begin{figure*}[t]
\centering
\begin{tabular}{ccccc ccccc}
% %%%%%%%%%%%%%% example 1 row 1
\begin{subfigure}[b]{0.12\textwidth}
\centering
\includegraphics[width=\linewidth]{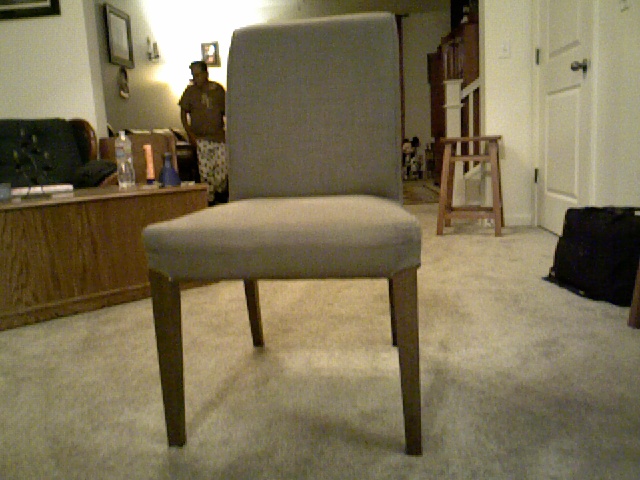}
\end{subfigure}
&

\multirow{2}{*}{
\begin{subfigure}[t]{0.065\textwidth}\vspace*{-0.7cm}
\centering
\includegraphics[width=\linewidth]{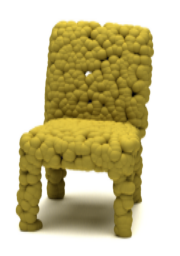}
\end{subfigure}}

                  &      
\begin{subfigure}[b]{0.065\textwidth}
\centering
\includegraphics[width=\linewidth]{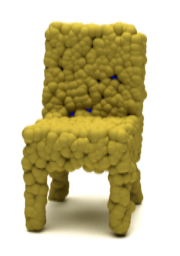}
\end{subfigure}
                  &     
\begin{subfigure}[b]{0.065\textwidth}
\centering
\includegraphics[width=\linewidth]{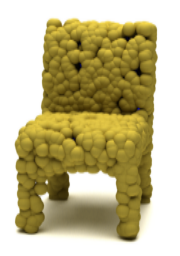}
\end{subfigure}
                  &    
\begin{subfigure}[b]{0.065\textwidth}
\centering
\includegraphics[width=\linewidth]{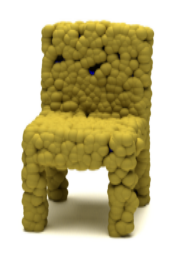}
\end{subfigure} &
%%%%%%%%%%%%%%%%%%%%%%%%%%%
%%%%%%%%%%%%%% example 2 row 1
\begin{subfigure}[b]{0.12\textwidth}
\centering
\includegraphics[width=\linewidth]{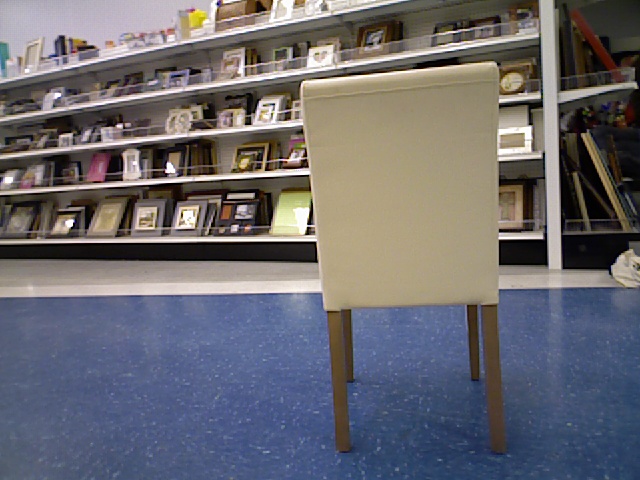}
\end{subfigure}
&

\multirow{2}{*}{
\begin{subfigure}[t]{0.065\textwidth}\vspace*{-0.7cm}
\centering
\includegraphics[width=\linewidth]{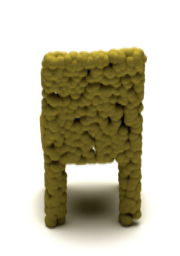}
\end{subfigure}}

                  &      
\begin{subfigure}[b]{0.065\textwidth}
\centering
\includegraphics[width=\linewidth]{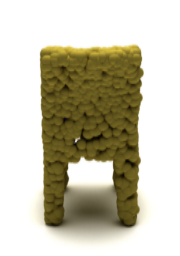}
\end{subfigure}
                  &     
\begin{subfigure}[b]{0.065\textwidth}
\centering
\includegraphics[width=\linewidth]{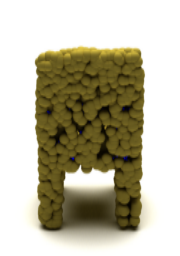}
\end{subfigure}
                  &    
\begin{subfigure}[b]{0.065\textwidth}
\centering
\includegraphics[width=\linewidth]{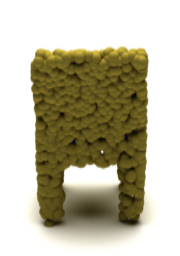}
\end{subfigure}
%%%%%%%%%%%%%%%%%%%%%%%%%%%%
%%%%%%%%%%%%%% example 1 row 2
                  \\
\begin{subfigure}[b]{0.12\textwidth}
\centering
\includegraphics[width=\linewidth]{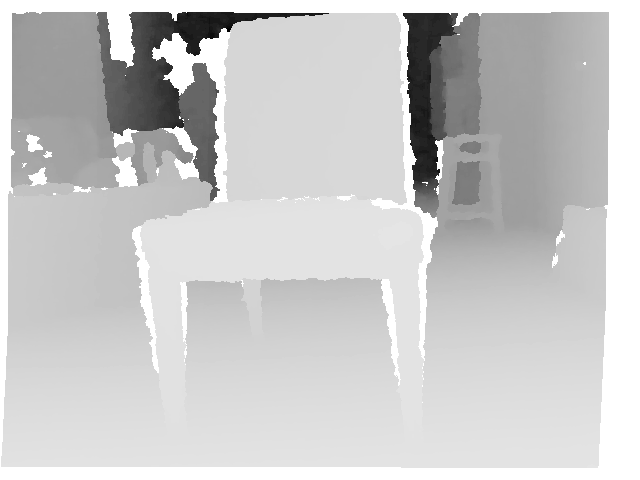}
\end{subfigure}
                  &  
                   
                  &      
\begin{subfigure}[b]{0.065\textwidth}
\centering
\includegraphics[width=\linewidth]{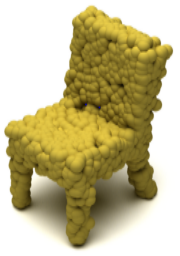}
\end{subfigure}
                  &     
\begin{subfigure}[b]{0.065\textwidth}
\centering
\includegraphics[width=\linewidth]{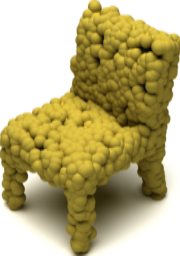}
\end{subfigure}
                  &    
\begin{subfigure}[b]{0.065\textwidth}
\centering
\includegraphics[width=\linewidth]{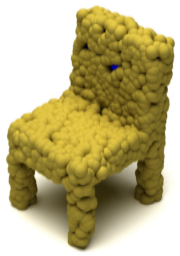}
\end{subfigure} &
%%%%%%%%%%%%%%%%%%%%%%%%%%%%
%%%%%%%%%%%%%% example 2 row 2
                  
\begin{subfigure}[b]{0.12\textwidth}
\centering
\includegraphics[width=\linewidth]{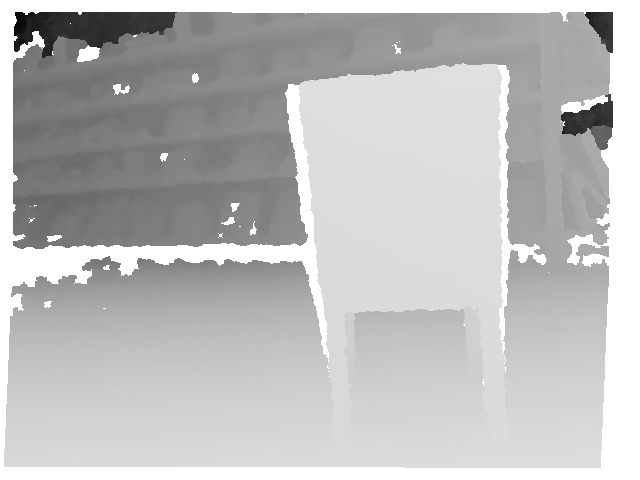}
\end{subfigure}
                  &  
                   
                  &      
\begin{subfigure}[b]{0.065\textwidth}
\centering
\includegraphics[width=\linewidth]{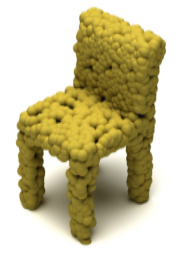}
\end{subfigure}
                  &     
\begin{subfigure}[b]{0.065\textwidth}
\centering
\includegraphics[width=\linewidth]{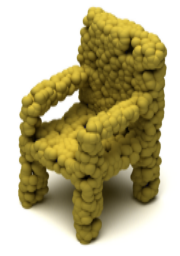}
\end{subfigure}
                  &    
\begin{subfigure}[b]{0.065\textwidth}
\centering
\includegraphics[width=\linewidth]{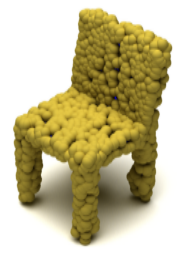}
\end{subfigure}
                  \\
%%%%%%%%%%%%%%%%%%%%%%%%%%%
%%%%%%%%%%%%%% example 3 row 1
 \begin{subfigure}[b]{0.12\textwidth}
\centering
\includegraphics[width=\linewidth]{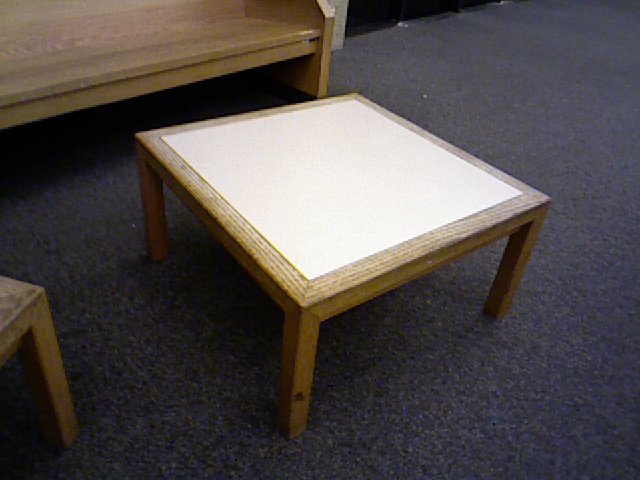}
\end{subfigure}
&

\multirow{2}{*}{
\begin{subfigure}[t]{0.065\textwidth}\vspace*{-0.7cm}
\centering
\includegraphics[width=\linewidth]{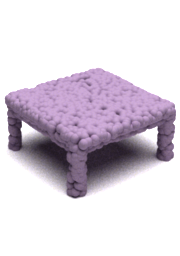}
\end{subfigure}}

                  &      
\begin{subfigure}[b]{0.065\textwidth}
\centering
\includegraphics[width=\linewidth]{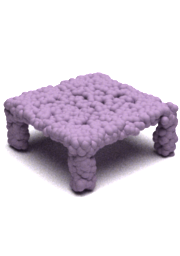}
\end{subfigure}
                  &     
\begin{subfigure}[b]{0.065\textwidth}
\centering
\includegraphics[width=\linewidth]{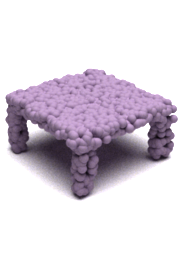}
\end{subfigure}
                  &    
\begin{subfigure}[b]{0.065\textwidth}
\centering
\includegraphics[width=\linewidth]{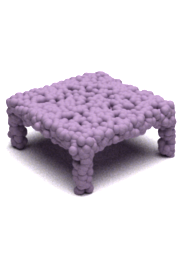}
\end{subfigure} &
%%%%%%%%%%%%%%%%%%%%%%%%%%%
%%%%%%%%%%%%%% example 4 row 1
\begin{subfigure}[b]{0.12\textwidth}
\centering
\includegraphics[width=\linewidth]{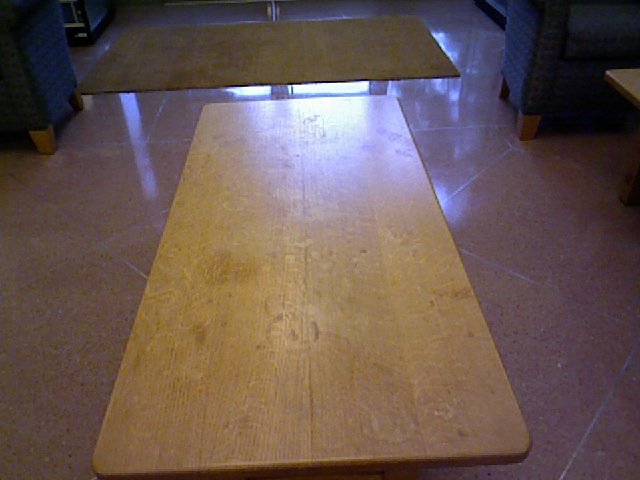}
\end{subfigure}
&

\multirow{2}{*}{
\begin{subfigure}[t]{0.065\textwidth}\vspace*{-0.7cm}
\centering
\includegraphics[width=\linewidth]{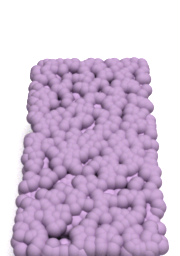}
\end{subfigure}}

                  &      
\begin{subfigure}[b]{0.065\textwidth}
\centering
\includegraphics[width=\linewidth]{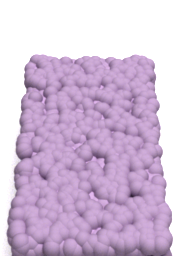}
\end{subfigure}
                  &     
\begin{subfigure}[b]{0.065\textwidth}
\centering
\includegraphics[width=\linewidth]{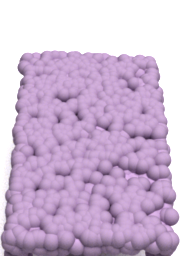}
\end{subfigure}
                  &    
\begin{subfigure}[b]{0.065\textwidth}
\centering
\includegraphics[width=\linewidth]{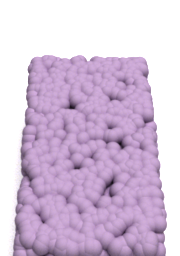}
\end{subfigure}
%%%%%%%%%%%%%%%%%%%%%%%%%%%%
%%%%%%%%%%%%%% example 3 row 2
                  \\
\begin{subfigure}[b]{0.12\textwidth}
\centering
\includegraphics[width=\linewidth]{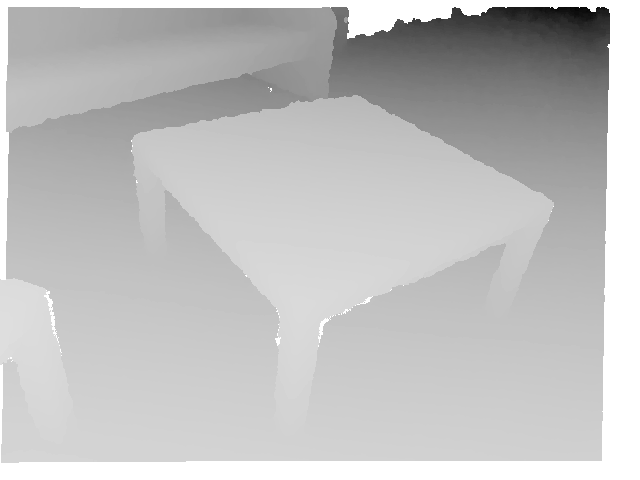}
\end{subfigure}
                  &  
                   
                  &      
\begin{subfigure}[b]{0.065\textwidth}
\centering
\includegraphics[width=\linewidth]{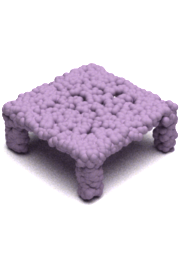}
\end{subfigure}
                  &     
\begin{subfigure}[b]{0.065\textwidth}
\centering
\includegraphics[width=\linewidth]{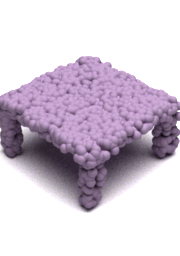}
\end{subfigure}
                  &    
\begin{subfigure}[b]{0.065\textwidth}
\centering
\includegraphics[width=\linewidth]{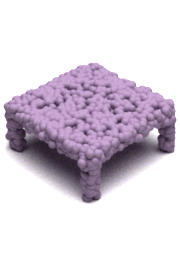}
\end{subfigure} &
%%%%%%%%%%%%%%%%%%%%%%%%%%%%
%%%%%%%%%%%%%% example 3 row 2
                  
\begin{subfigure}[b]{0.12\textwidth}
\centering
\includegraphics[width=\linewidth]{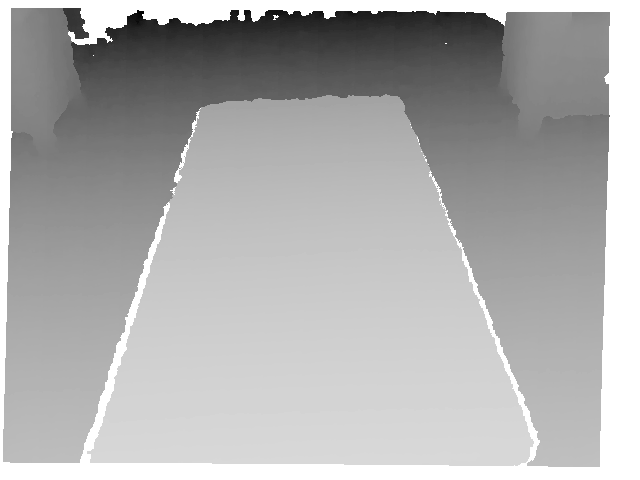}
\end{subfigure}
                  &  
                   
                  &      
\begin{subfigure}[b]{0.065\textwidth}
\centering
\includegraphics[width=\linewidth]{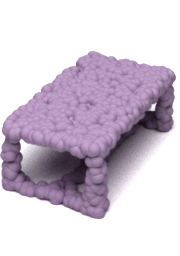}
\end{subfigure}
                  &     
\begin{subfigure}[b]{0.065\textwidth}
\centering
\includegraphics[width=\linewidth]{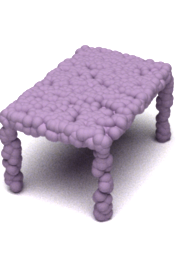}
\end{subfigure}
                  &    
\begin{subfigure}[b]{0.065\textwidth}
\centering
\includegraphics[width=\linewidth]{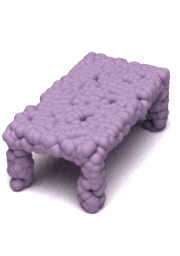}
\end{subfigure}\\

    RGB-D & GT & \multicolumn{3}{>{\centering\arraybackslash}c}{Completion Results} & RGB-D & GT & \multicolumn{3}{>{\centering\arraybackslash}c}{Completion Results}\\

\end{tabular}
\caption{Application of our model on scans from the Redwood 3DScans dataset. \model takes partial point clouds induced from the depth maps, not the RGB image as input. Left: from a more complete view, the model outputs stable, similar completions. Right: from an uncertain viewpoint, the model outputs multiple completions with a larger variation.}
\label{fig:real_mm}
%\vspace{-10pt}
\end{figure*}

\myparagraph{Results on ShapeNet.} The shape completion model trained in Section~\ref{sect:shape_complete} is directly used to demonstrate completion diversity on ShapeNet, as shown in Figure~\ref{fig:mm_shapenet}. We choose a bottom view of a chair and show that, in the top row, all of our completion results match well with the constrained viewpoint, and in the bottom row, our completion results are noticeably diverse from the canonical viewpoint.

\myparagraph{Results on real scans.} 
%As PartNet and ShapeNet are based on given meshes, it can be different from real objects and real scans. Therefore, 
We further investigate how our model pre-trained on ShapeNet can perform on scans of real objects. We use the Redwood 3DScans dataset~\cite{choi2016large} and test our model on partial shapes of chairs and tables, sampled from its depth images. Since the GenRe benchmark~\cite{zhang2018learning} does not provide table data, the training data for the table category are generated by randomly sampling 20 views from ShapeNet meshes, following GenRe's procedure. Within each example, we present the real RGB-D scans and ground-truths from the input views. 

Figure~\ref{fig:real_mm} shows results on two views of two different chairs and tables. For chair scans, the left example shows the front view, and the completion results only vary slightly across different runs. The right chair example shows a back view, and the uncertainty allows more varied completion. Similarly, the left table scan shows a large part of the table, so the completion varies less than the right example, which only shows the top of the table.

\section{Conclusion}
We have introduced \model, a unified framework for both shape generation and shape completion. Our model, trained on a simple $\mathcal{L}_2$ loss, is based on diffusion probabilistic models and learns to reverse a diffusion process by progressively removing noise from noise-initialized samples. A minor modification on the objective also results in a shape completion model without the need for any architectural change. Experimentally, we show the failure with straight-forward extension of diffusion models to either pure voxel or point representations. With the point-voxel representation, our model demonstrates superior generative power and impressive shape completion quality. Unlike most baseline models which use deterministic encoder-decoder structures, \model can output multiple possible completion results given a partial shape. Additionally, it can complete real 3D scans, thus offering practical usage in various downstream applications.

\myparagraph{Acknowledgements:} Yilun Du is funded by an NSF graduate fellowship. This work is in part supported by ARMY MURI grant W911NF-15-1-0479, Samsung Global Research Outreach (GRO) program, Amazon Research Award (ARA), IBM, Stanford HAI, and Autodesk.

{\small
\bibliographystyle{ieee_fullname}
\bibliography{egbib}
}

\newpage

\newcommand{\appendixhead}%
{\centering\textbf{\Large Appendix: 3D Shape Generation and Completion through Point-Voxel Diffusion}
\vspace{0.25in}}

\twocolumn[\appendixhead]

\appendix

\section{Additional Generation Metrics}
We present additional generation metrics in Table~\ref{tab:gen_full_metrics}, following PointFlow~\cite{yang2019pointflow}. We report coverage (COV), which measures the fraction of point clouds in the reference set that are matched to at least one point cloud in the generated set. We further report minimum matching distance (MMD), which measures for each point cloud in a reference set, the distance to its nearest neighbor in the generated set. Note that these generation metrics can vary depending on implementation and do not necessarily correlate to generation quality, as discussed in~\cite{yang2019pointflow}. 

Table~\ref{tab:gen-vox} includes generation results on Airplane, Chair, Car compared with the voxel-diffusion model, Vox-Diff, as described in the main paper, evaluated using the 1-NN metric. By generating less noisy point clouds, \model significantly outperforms Vox-Diff. 

\begin{table*}[h]
\centering
\small
\begin{tabular}{lcccccccccccc}
\toprule
\multirow{4}{*}{Model} &
& \multicolumn{2}{c}{Airplane} & \multicolumn{6}{c}{Chair} & \multicolumn{2}{c}{Car} \\
\cmidrule(lr){2-5}\cmidrule(lr){6-9}\cmidrule(lr){10-13}
& \multicolumn{2}{c}{MMD$\downarrow$} &
\multicolumn{2}{c}{COV$\uparrow$ (\%)} & \multicolumn{2}{c}{MMD$\downarrow$}&
\multicolumn{2}{c}{COV$\uparrow$ (\%)} & \multicolumn{2}{c}{MMD$\downarrow$}&
\multicolumn{2}{c}{COV$\uparrow$ (\%)} \\ 
\cmidrule(lr){2-3} \cmidrule(lr){4-5}\cmidrule(lr){6-7} \cmidrule(lr){8-9}  
\cmidrule(lr){10-11} \cmidrule(lr){12-13} 
         &    CD   &     EMD &  CD    &     EMD   &    CD   &     EMD &  CD    &     EMD   &    CD   &     EMD &  CD    &     EMD   \\ 
\midrule
                r-GAN~\cite{achlioptas2018learning}   & 0.4471 & 2.309 & 30.12 & 14.32 & 5.151 & 8.312 & 24.27 & 15.13 & 1.446 & 2.133 & 19.03 & 6.539\\
                l-GAN (CD)~\cite{achlioptas2018learning}  & 0.3398 & 0.5832 & 38.52 & 21.23 & 2.589 & 2.007 & 41.99 & 29.31 & 1.532 & 1.226 & 38.92 & 23.58\\
                l-GAN (EMD)~\cite{achlioptas2018learning} & 0.3967 & 0.4165 & 38.27 & 38.52 & 2.811 & 1.619 & 38.07 & 44.86 & 1.408 & 0.8987 & 37.78 & 45.17\\
                  PointFlow~\cite{yang2019pointflow}    &0.2243 &0.3901 &47.90 &46.41 & \textbf{2.409}&1.595 &42.90 &50.00 & \textbf{0.9010}& 0.8071& 46.88&50.00 \\ 
                  SoftFlow~\cite{kim2020softflow}     & 0.2309 & \textbf{0.3745} &46.91 &47.90 & 2.528 & 1.682 & 41.39 & 47.43 & 1.187 & 0.8594 & 42.90 & 44.60 \\ 
                   DPF-Net~\cite{klokov2020discrete}  &0.2642 &0.4086 &46.17 & 48.89 & 2.536&1.632 &44.71 &48.79 &1.129 & 0.8529& 45.74&49.43 \\ 
                  Shape-GF~\cite{cai2020learning}  & 2.703 & 0.6592 & 40.74 &40.49 & 2.889 & 1.702 & 46.67 & 48.03 & 9.232 & \textbf{0.7558} &\textbf{49.43} & 50.28\\
                   Vox-Diff  & 1.322 &0.5610 &11.82 &25.43 & 5.840 & 2.930 &17.52 & 21.75 & 5.646 & 1.551 & 6.530 & 22.15 \\ 
                  \model (Ours) &\textbf{0.2243} &0.3803 &\textbf{48.88} &\textbf{52.09} &2.622&\textbf{1.556} &\textbf{49.84} & \textbf{50.60} &1.077 & 0.7938 &41.19 &\textbf{50.56} \\ 
\bottomrule
\end{tabular}

\caption{Additional generation metrics, following PointFlow~\cite{yang2019pointflow}.}
\label{tab:gen_full_metrics}
\end{table*}

\begin{table}[t]
\centering
\small
\setlength{\tabcolsep}{4pt}
\begin{tabular}{lcccccc}
\toprule
            & \multicolumn{2}{c}{Airplane}             & \multicolumn{2}{c}{Chair}                 & \multicolumn{2}{c}{Car}         \\ 
\cmidrule(lr){2-3}\cmidrule(lr){4-5}\cmidrule(lr){6-7}
            % & \multicolumn{2}{c|}{1-NN, \%}             & \multicolumn{2}{c|}{1-NN, \%}             & \multicolumn{2}{c}{1-NN, \%}    \\ \cline{2-7} 
            & CD             & \multicolumn{1}{c}{EMD} & CD             & \multicolumn{1}{c}{EMD} & CD             & EMD            \\ \midrule
Vox-Diff   & 99.75         & 98.13                    & 97.12          & 96.74                    & 99.56          & 96.83          \\
\model (ours)   & \textbf{73.82} & \textbf{64.81}           & \textbf{56.26} & \textbf{53.32}           & \textbf{54.55} & \textbf{53.83}\\
\bottomrule
\end{tabular}
\caption{Generation results on Airplane, Chair, Car compared with the voxel-diffusion model, Vox-Diff, as described in the main paper, evaluated using the 1-NN metric. By generating less noisy point clouds, \model significantly outperforms Vox-Diff. }
\label{tab:gen-vox}
\end{table}

\section{Point Cloud Generation Visualization}
\begin{figure*}[p!]
    \centering
    \includegraphics[width=\textwidth]{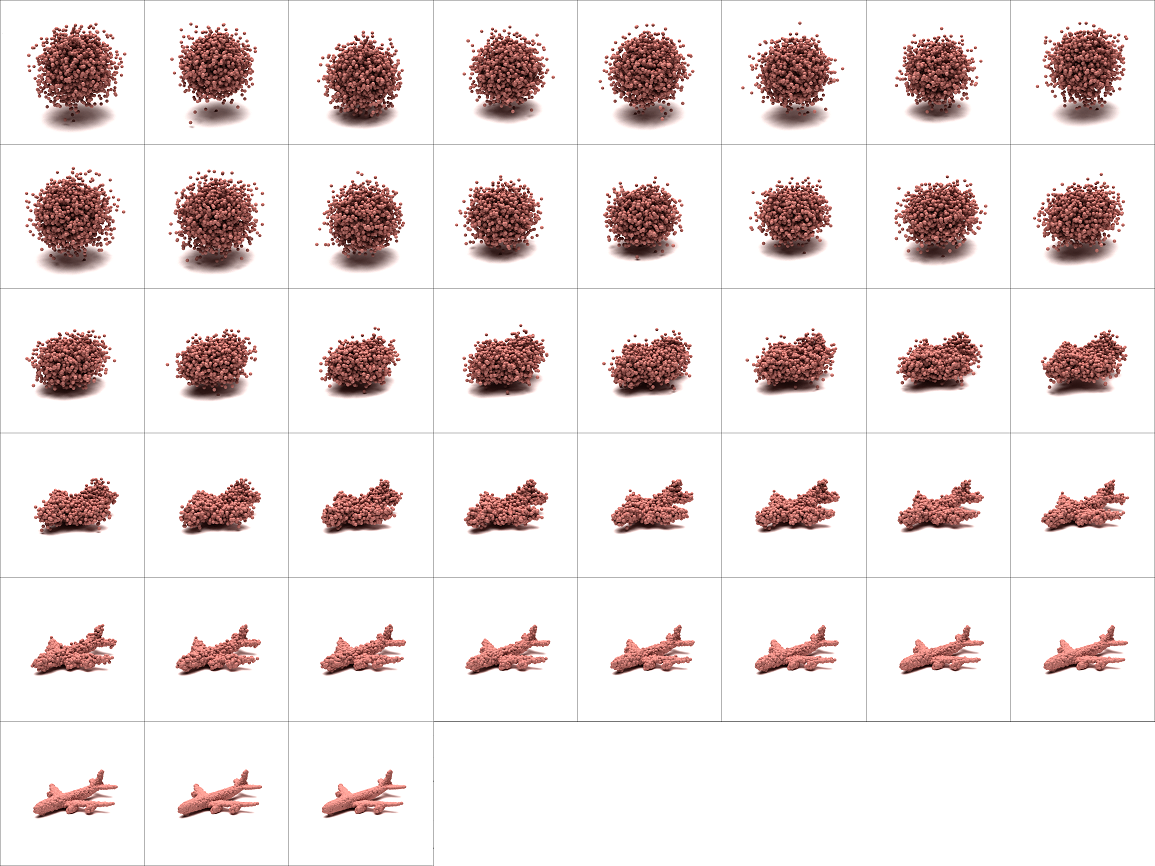}
\caption{Airplane generation process.}
\label{fig:airplane-gen-process}
\end{figure*}

\begin{figure*}[p]
    \centering
    \includegraphics[width=\textwidth]{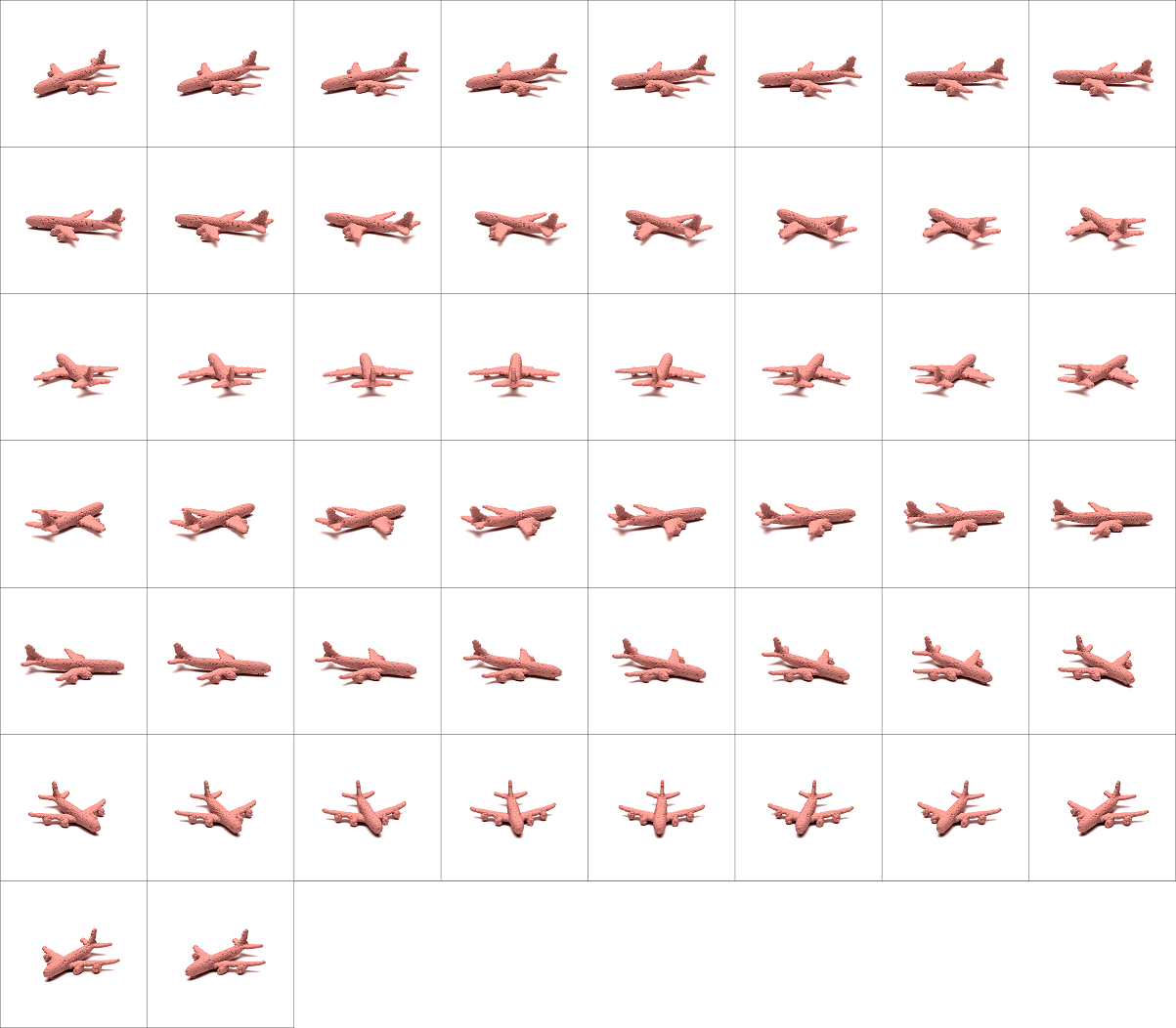}
\caption{Airplane results from all angles.}
\label{fig:airplane-rotate-process}
\end{figure*}

\begin{figure*}[p]
    \centering
    \includegraphics[width=\textwidth]{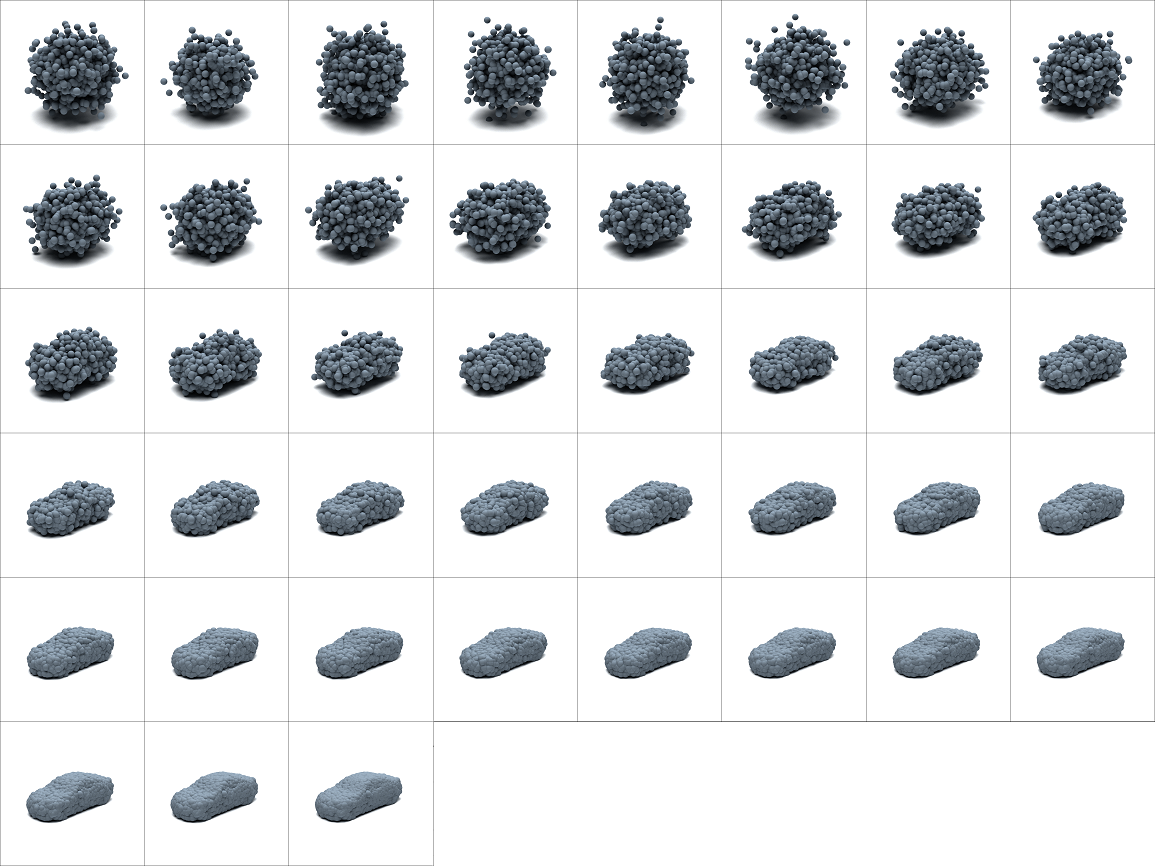}
\caption{Car generation process.}
\label{fig:car-gen-process}
\end{figure*}

\begin{figure*}[p]
    \centering
    \includegraphics[width=\textwidth]{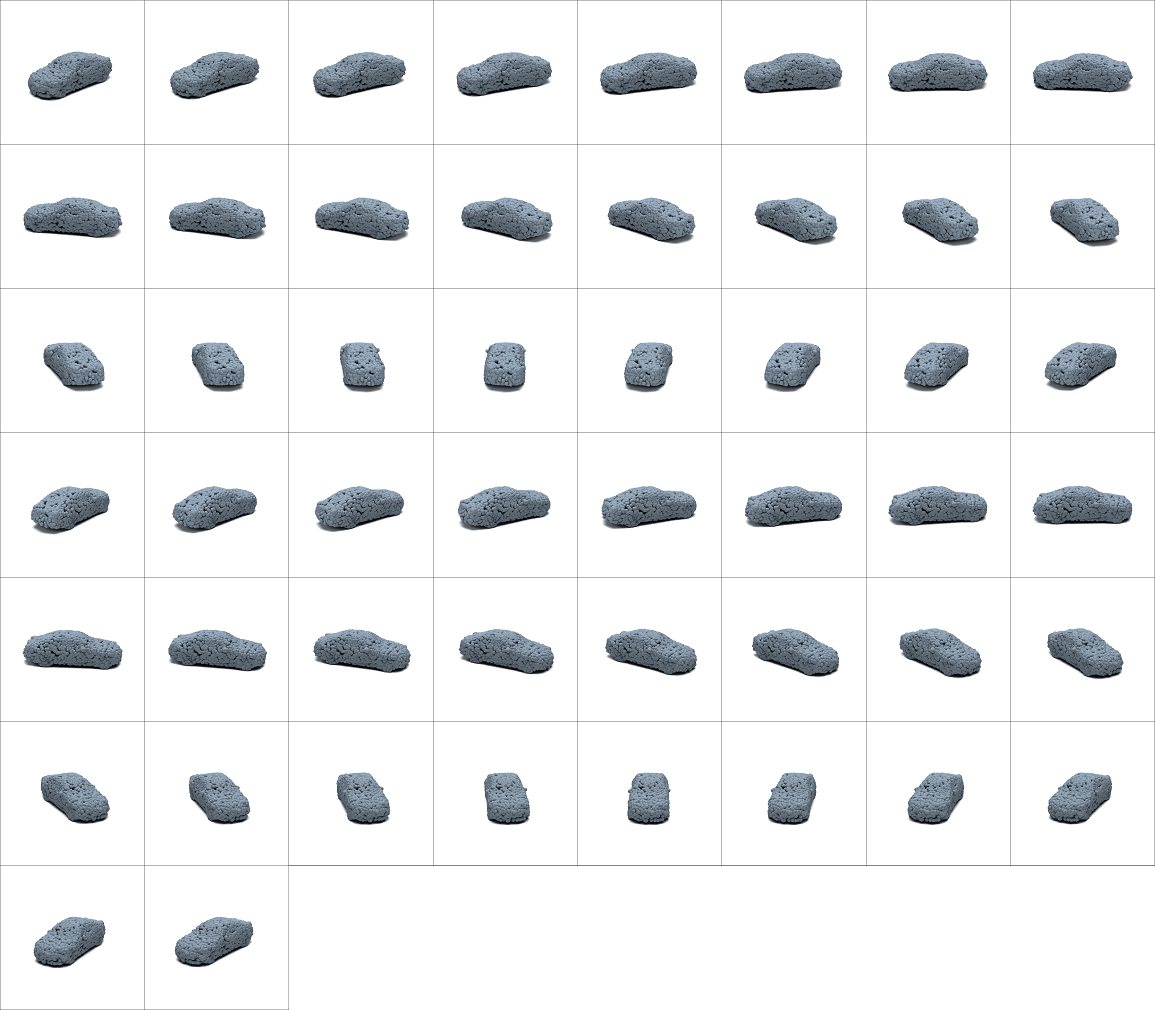}
\caption{Car results from all angles.}
\label{fig:car-rotate-process}
\end{figure*}

\begin{figure*}[p]
    \centering
    \includegraphics[width=\textwidth]{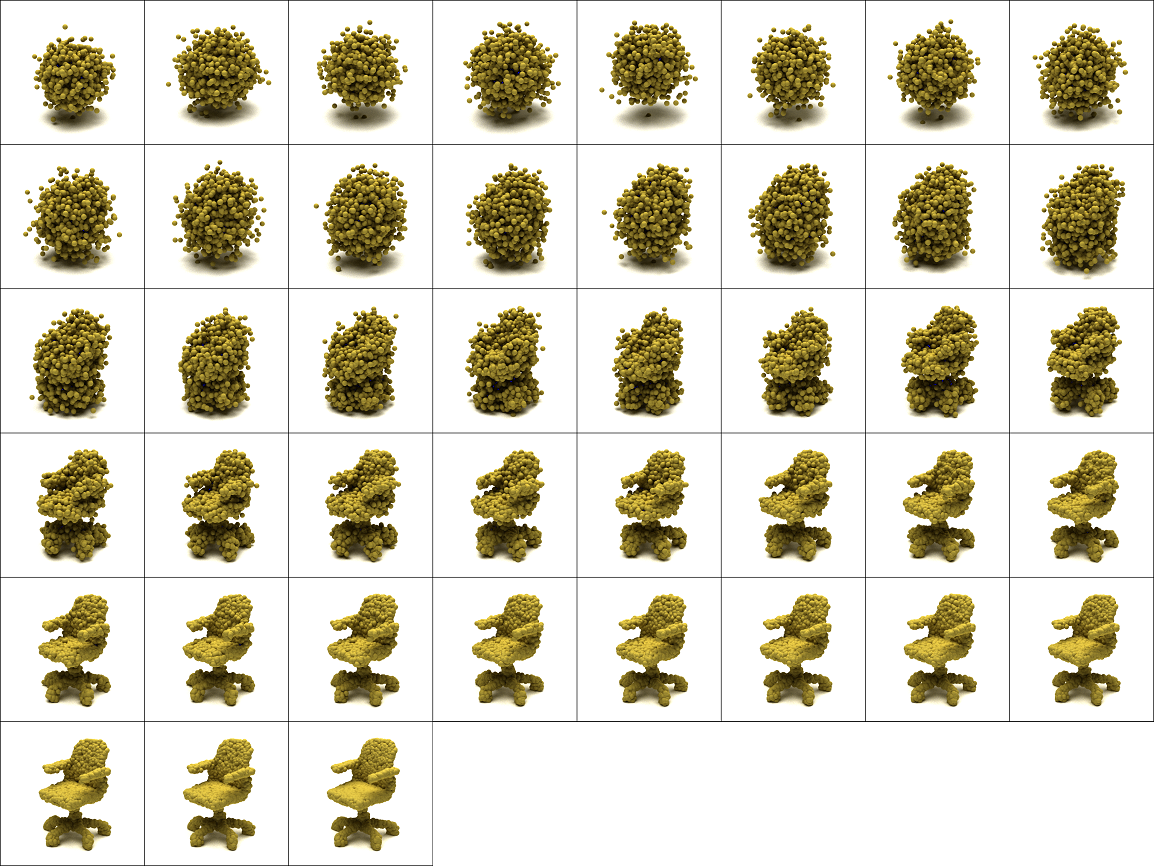}
\caption{Chair generation process.}
\label{fig:chair-gen-process}
\end{figure*}

\begin{figure*}[p]
    \centering
    \includegraphics[width=\textwidth]{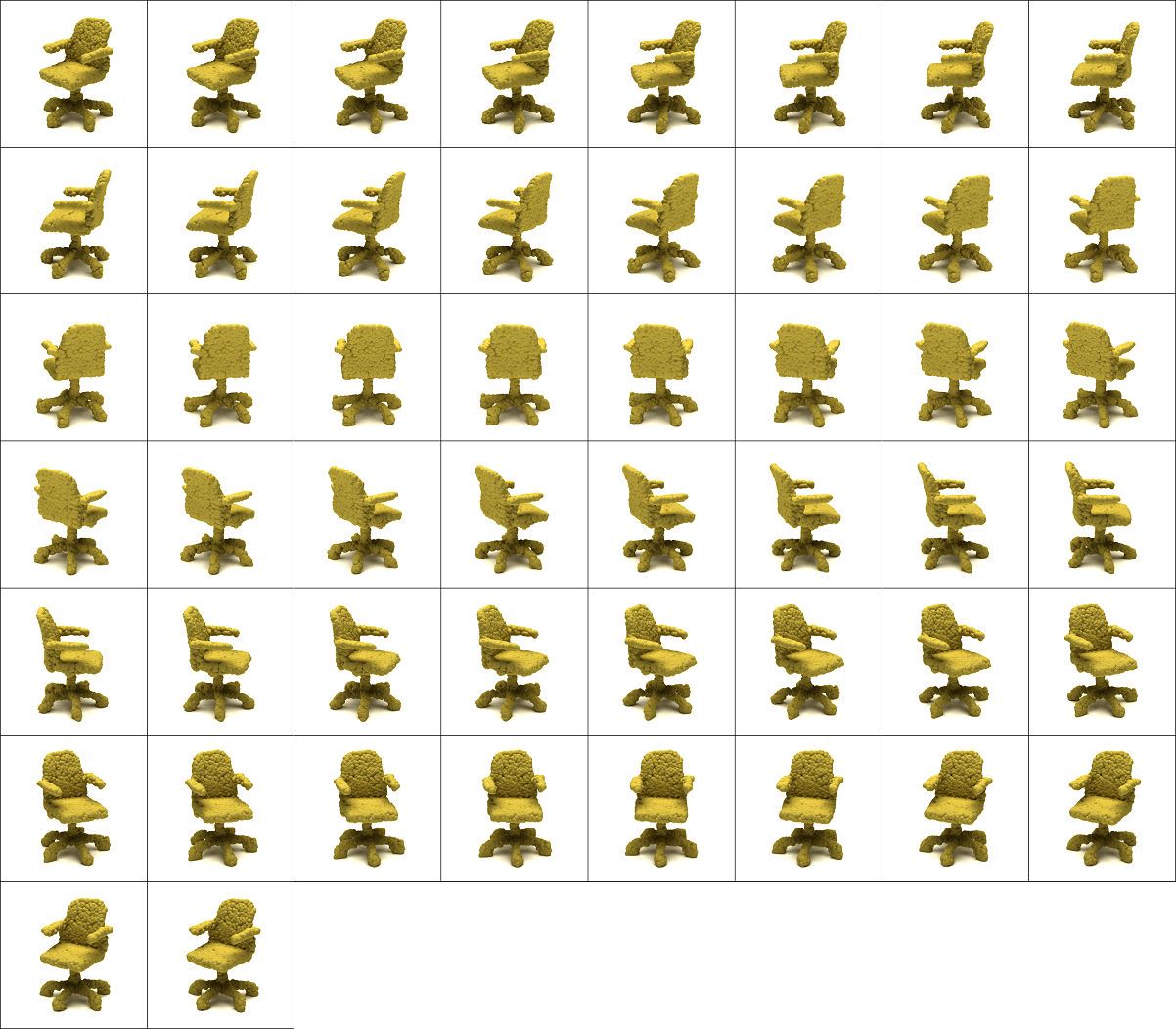}
\caption{Chair results from all angles.}
\label{fig:chair-rotate-process}
\end{figure*}

\begin{figure*}[p]
    \centering
    \includegraphics[width=\textwidth]{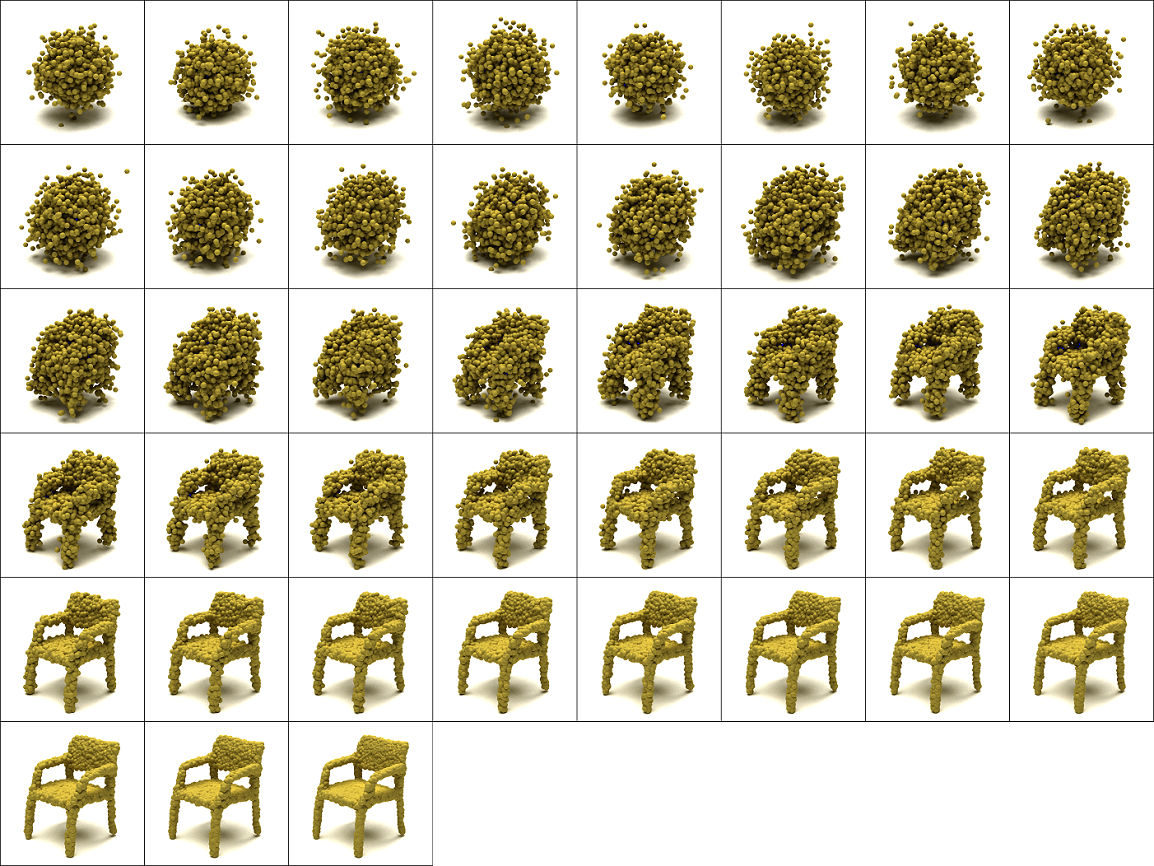}
\caption{Chair generation process.}
\label{fig:chair1-gen-process}
\end{figure*}

\begin{figure*}[p]
    \centering
    \includegraphics[width=\textwidth]{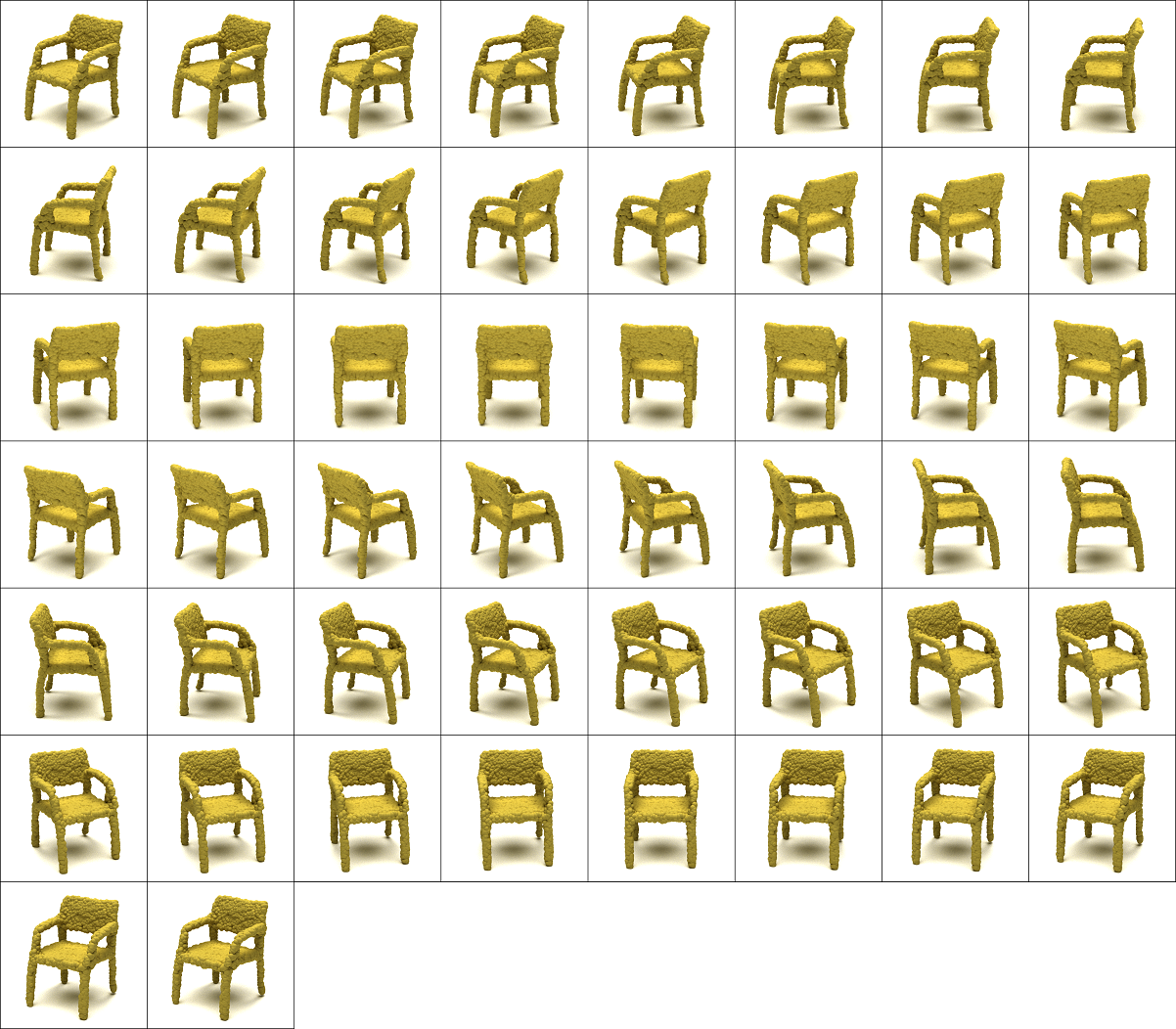}
\caption{Chair results from all angles.}
\label{fig:chair1-rotate-process}
\end{figure*}

We additionally visualize some generation results for Airplane, Car, and Chair in terms of the generation process and the final generated shapes from all angles in Figures~\ref{fig:airplane-gen-process},~\ref{fig:airplane-rotate-process},~\ref{fig:car-gen-process}.~\ref{fig:car-rotate-process},~\ref{fig:chair-gen-process},~\ref{fig:chair-rotate-process},~\ref{fig:chair1-gen-process}, and \ref{fig:chair1-rotate-process}.

\section{Derivation of the Variational Lower Bound}

\begin{equation}
\begin{aligned}
    \E_{q(\x_0)}[\log p_\theta(\x_0)] 
    &= \E_{q(\x_0)}\Big[\log \int p_\theta(\x_0,...,\x_T) \,d\x_{1:T}\Big] \\
    &\geq  \E_{q(\x_0)}\Big[\int q(\x_1,...,\x_T|\x_0) \\
    &\log \frac{p_\theta(\x_0,...,\x_T)}{q(\x_1,...,\x_T|\x_0)} \,d\x_{1:T}\Big]\\
    &= \E_{q(\x_{0:T})}\Big[\log \frac{p_\theta(\x_0,...,\x_T)}{q(\x_1,...,\x_T|\x_0)}\Big]\nonumber,
\end{aligned}
\end{equation}
where the inequality is by Jensen's inequality.

\section{Properties of the Diffusion Model}
$\{\beta_0,...,\beta_T\}$ is a sequence of increasing parameters; $\alpha_t = 1 - \beta_t$ and $\tilde{\alpha}_t = \prod_{s=1}^{t} \alpha_s$. Two following properties are crucial to deriving the final $\mathcal{L}_2$ loss.

\textbf{Property 1.} Tractable marginal of the forward process:
\begin{equation}
\begin{aligned}
q(\x_t | \x_0) &= \int q(\x_{1:t}|\x_0) \,d\x_{1:{(t-1)}}\\
&= \N(\sqrt{\tilde{\alpha}_t} \x_0, (1-\tilde{\alpha}_t) \I) \nonumber .
\end{aligned}
\end{equation}
This property is proved in the Appendix of~\cite{ho2020denoising} and provides convenient closed-form evaluation of $\x_t$ knowing $\x_0$:
\begin{equation}\label{eq:xt-from-x0}
    \x_t = \sqrt{\tilde{\alpha}_t} \x_0 + \sqrt{1-\tilde{\alpha}_t}\bm{\epsilon},
\end{equation}
where $\bm{\epsilon} \sim \N(0, \I)$.

\textbf{Property 2.} Tractable posterior of the forward process.
We first note the Bayes' rule that connects the posterior with the forward process,
\[
q(\x_{t-1} | \x_t, \x_0) = \frac{q(\x_t | \x_{t-1}, \x_0)q(\x_{t-1}|\x_0)}{q(\x_t |\x_0)}. 
\]
Since the three probabilities on the right are Gaussian, the posterior is also Gaussian, given by
\begin{equation}\label{eq:posterior-appendix}
\begin{aligned}
&q(\x_{t-1} | \x_t, \x_0) = \\
&\N(\frac{\sqrt{\tilde{\alpha}_{t-1}} \beta_t}{1-\tilde{\alpha}_t} \x_0 +\frac{\sqrt{\alpha_t} (1-\tilde{\alpha}_{t-1})}{1-\tilde{\alpha}_t} \x_t,
\frac{(1-\tilde{\alpha}_{t-1})}{1-\tilde{\alpha}_t}\beta_t \I).
\end{aligned}
\end{equation}

\section{Derivation of $\mathcal{L}_2$ Loss}

We need to match generative transition $p_\theta(\x_{t-1}|\x_t)$ with ground-truth posterior $q(\x_{t-1} | \x_t, \x_0)$, both of which are Gaussian with a pre-determined variance schedule $\beta_1, ..., \beta_T$. Therefore, maximum likelihood learning is reduced to simple $\mathcal{L}_2$ loss of the form with two cases:
\begin{align*}
    L_t = 
    \begin{cases}
    \norm{\frac{\sqrt{\tilde{\alpha}_{t-1}} \beta_t}{1-\tilde{\alpha}_t} \x_0 +\frac{\sqrt{\alpha_t} (1-\tilde{\alpha}_{t-1})}{1-\tilde{\alpha}_t} \x_t - \mu_\theta(\x_t, t)}^2, & t>1\\[10pt]
    \norm{\x_0 - \mu_\theta(\x_t, t)}^2, & t=1
    \end{cases}
\end{align*}
where $\alpha_t = 1 - \beta_t$ and $\tilde{\alpha}_t = \prod_{s=1}^{t} \alpha_s$. The supervision target of case $t>1$ comes from Eqn.~\ref{eq:posterior-appendix}. We can Further reduce the case when $t>1$ by substituting $\x_0$ as an expression of $\x_t$ using Eqn.~\ref{eq:xt-from-x0} and arrive at
\begin{equation}\label{eq:l2-obj}
    \mathcal{L}_t = 
    \begin{cases}
    \norm{\frac{1}{\sqrt{\alpha_t}}\left(\x_t - \frac{\beta_t}{\sqrt{1-\tilde{\alpha}_t}} \bm{\epsilon}\right) - \mu_\theta(\x_t, t)}^2, & t>1\\[10pt]
    \norm{\x_0 - \mu_\theta(\x_t, t)}^2, & t=1
    \end{cases}\nonumber
\end{equation}
where $\bm{\epsilon} \sim \N(0, \I)$.

Note that when $t=1$, $\tilde{\alpha}_1 = \alpha_1$ so that the supervision target of the first case above evaluated at $t=1$ becomes:
\begin{equation}
    \begin{aligned}
    \frac{1}{\sqrt{\alpha_1}}\left(\x_1 - \frac{\beta_t}{\sqrt{1-\tilde{\alpha}_1}} \bm{\epsilon}\right) &= \frac{1}{\sqrt{\tilde{\alpha}_1}}\left(\x_1 - \sqrt{1-\tilde{\alpha}_1} \bm{\epsilon}\right)\\
    &= \x_0,
    \end{aligned}
\end{equation}
where the last equality is by rewriting Eqn.~\ref{eq:xt-from-x0}. Therefore, in fact, the two cases are equivalent.

The final $\mathcal{L}_2$ loss is 
\begin{equation}
    \mathcal{L}_t = 
    \norm{\frac{1}{\sqrt{\alpha_t}}\left(\x_t - \frac{\beta_t}{\sqrt{1-\tilde{\alpha}_t}} \bm{\epsilon}\right) - \mu_\theta(\x_t, t)}^2.
   \nonumber
 \end{equation}
Since $\x_t$ is known when $\x_0$ is known, we can redefine the model output as $\bm{\epsilon}_\theta(\x_t, t)$.
Instead of directly predicting $\mu_\theta(\x_t, t)$, we instead predict a noise value $\epsilon_\theta(\x_t, t)$, where 
\begin{equation}\label{eq:p_mean}
    \mu_\theta(\x_t, t) = \frac{1}{\sqrt{\alpha_t}}\left(\x_t - \frac{\beta_t}{\sqrt{1-\tilde{\alpha}_t}} \bm{\epsilon}_\theta(\x_t, t)\right).
\end{equation}

Substituting in $\mu_\theta(\x_t, t)$ into the loss, we can arrive at the final loss
\begin{equation}\label{eq:l2-eps-appendix}
    \norm{\bm{\epsilon} - \bm{\epsilon}_\theta(\x_t, t)}^2,\;\; \bm{\epsilon} \sim \N(0, \I).
\end{equation}

\section{Point Cloud Generation Process}

Since the transition mean $\mu_\theta(\x_t, t)$ of $p_\theta(\x_{t-1}|\x_t)$ is calculated by Eqn.~\ref{eq:p_mean}, the generative process is performed by progressively sampling from $p_\theta(\x_{t-1}|\x_t)$ as $t=T,...,1$:
\begin{equation}\label{eq:generate-appendix}
    \x_{t-1} = \frac{1}{\sqrt{\alpha_t}}\left(\x_t - \frac{1 - \alpha_t}{\sqrt{1-\tilde{\alpha}_t}} \bm{\epsilon}_\theta(\x_t, t)\right) + \sqrt{\beta_t} \mathbf{z},
\end{equation}
where $\mathbf{z} \sim \N(0, \I)$. This approach is also similar to Langevin dynamics~\cite{du2019implicit, nijkamp2019learning} used in energy-based models, since it similarly adds scaled noise outputs from the model to current samples. Specifically, Langevin dynamics is in the following form:
\begin{equation}
    \label{eq:langevin}
\x_{t+1} = \x_t + s \frac{\partial}{\partial \x}\log p_\theta(\x) + \sqrt{2s}\mathbf{z},
\end{equation}
where $s$ denotes step size, $p_\theta(\x)$ denotes the model distribution, and $\mathbf{z} \sim \N(0,\I)$. Both processes are Markovian, shifting the previous output by a model-dependent term and a noise term. The scaled model output of our model can also be seen as an approximation of gradients of an energy function.

\section{Controlled Completion}
\begin{figure}[H]
    \centering
    \includegraphics[width=\linewidth]{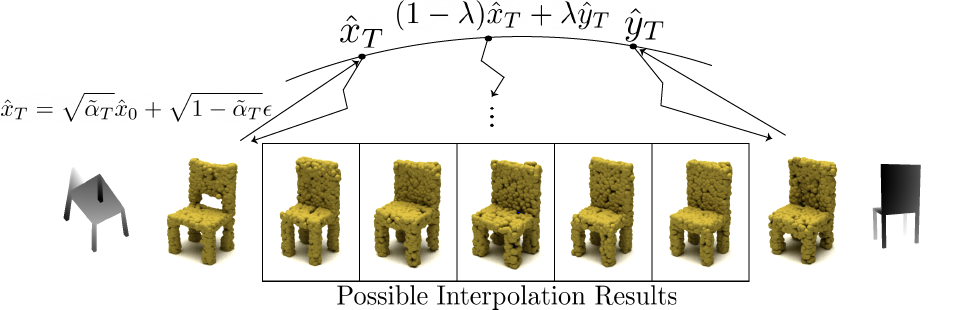}
\caption{Controlled completion process. }
\label{fig:control-gen}
\end{figure}

We show that our model can control the shape completion process in Figure~\ref{fig:control-gen}. Given a pretrained completion model, our formulation also enables control over completion results through latent interpolation. The figure below shows one such case: we may have used the left depth map to obtain a completed shape. But it comes with an unwanted feature (cavity at the chair's back). We can refine the result by feeding another depth map (shown on the right), with a better view of the back. To retain features from both partial shapes, we can interpolate by $(1-\lambda) \hat{\x}_T + \lambda\hat{\y}_T$ in the latent space at time $T$, where the latent features are obtained by $\hat{\x}_T = \sqrt{\tilde{\alpha}_T} \hat{\x}_0 + \sqrt{1-\tilde{\alpha}_T}\epsilon$ (see Eqn.~\ref{eq:xt-from-x0}). Interpolation presents diverse choices and users can actively control how much features are shared by varying $\lambda$.

\section{Training Details}

\subsection{Model Architecture}
\begin{figure*}[h!]
    \centering
    \includegraphics[width=0.8\textwidth]{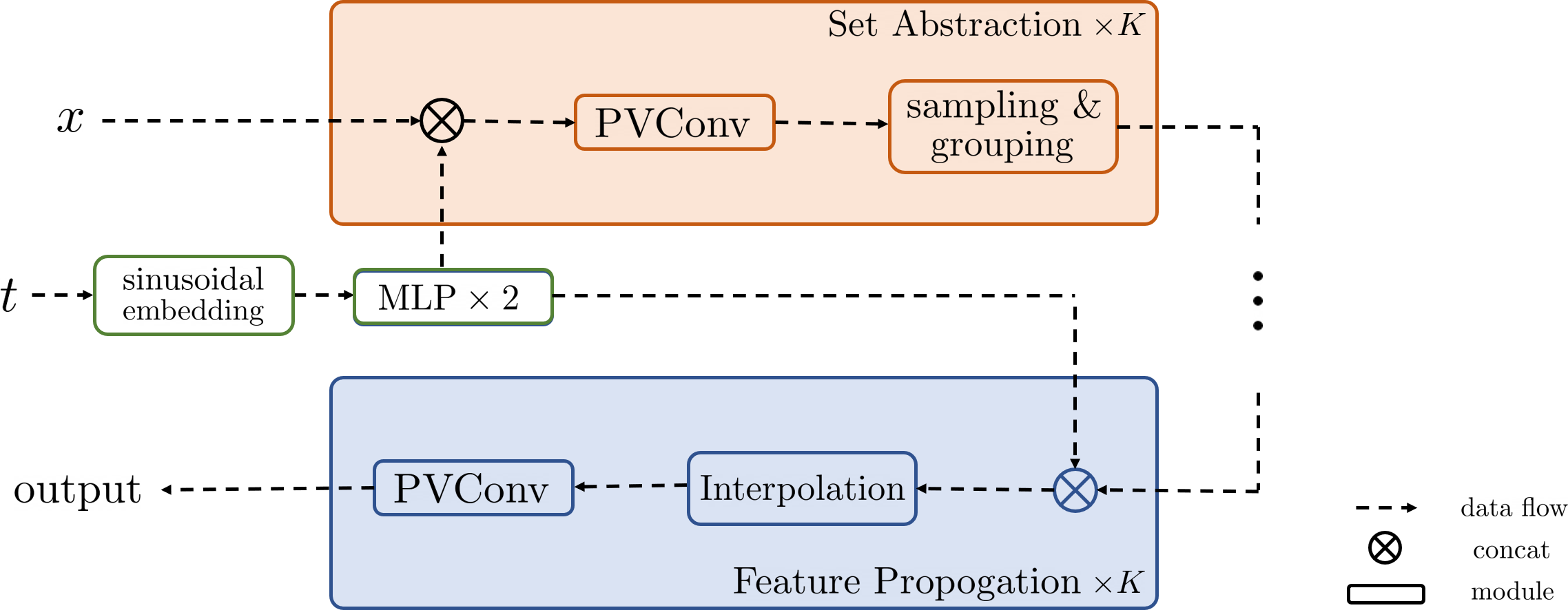}
\caption{Model architecture diagram. }
\label{fig:architecture}
\end{figure*}

Same as in~\cite{liu2019point}, our point-voxel CNN architecture is modified from PointNet++, where we replace the PointNet substructure with point-voxel convolution, as shown in Figure~\ref{fig:architecture}. We specify our architecture in Table~\ref{tab:sa}, Table~\ref{tab:fp}, and  and Table~\ref{tab:pvcnn}. Table~\ref{tab:sa} shows details of a single set abstraction (SA) module. Table~\ref{tab:fp} shows details of a single feature propagation (FP) module. Table~\ref{tab:pvcnn} shows how these modules are combined together.

In particular, we concatenate the temporal embeddings with point features before sending input into the Set Abstraction or the Feature Propagation modules. To obtain temporal embeddings, we used a sinusoidal positional embedding, commonly used in Transformers. Given a time $t$ and an embedding dimension $d$, the time embedding consists of pairs of $\sin$ and $\cos$ with varying frequencies,
$(\sin{(\omega_1 t)}, \cos{(\omega_1 t)}, ..., \sin{(\omega_{d/2} t)}, \cos{(\omega_{d/2} t)})$, where $\omega_k$ is $1/\left(10000^{2k/d}\right)$.

We use the same architecture for both generation and completion tasks. For shape completion specifically, the model takes as input a 200-point partial shape and 1,848 points sampled from noise, totaling 2048 points. At each step, the first 200 of the 2,048 points sampled by the model are replaced with the input partial shape. The updated point set is then used as input in the next time step.

\begin{table*}[h!]
\centering
\begin{tabular}{ |c|c|c| }
%\label{celeba}
\hline
\multicolumn{3}{|c|}{\bf{Set Abstraction}}\\
\hline
\multicolumn{3}{ |c| }{Input Feature Size: $N_{\text{input}} \times C_{\text{input}}$} \\
\hline
\multicolumn{3}{ |c| }{Input Time Embedding Size: $N_{\text{input}} \times E_t$} \\
\hline
\multicolumn{3}{ |c| }{Output Feature Size: $N_{\text{output}} \times C_{\text{output}}$} \\
\hline
\multicolumn{3}{ |c| }{Voxelization Resolution: $D$}\\
\hline
\multicolumn{3}{ |c| }{Number of Point-Voxel Convolution (PVConv) blocks: $L$}\\
\hline
\multicolumn{3}{ |c| }{Whether to use attention mechanism: use\_attn}\\
\hline
\hline
\multicolumn{3}{ |c| }{PVConv $\times L$} \\
\hline
Layers & In-Out Size & Stride \\ \hline
Input: (X, t) & ($D\times D\times D\times C_{\text{input}}$, $D\times D\times D\times E_t$) & \\
Concat(X, t) & -& - \\
3x3x3 conv($C_{\text{output}}$), GroupNorm(8), Swish & $D\times D\times D\times C_{\text{output}}$ & 1 \\
Dropout(0.1) & &\\
3x3x3 conv($C_{\text{output}}$), GroupNorm(8) & $D\times D\times D\times C_{\text{output}}$ & 1  \\
Attention(use\_attn) & $D\times D\times D\times C_{\text{output}}$ & \\
\hline
\hline
\multicolumn{3}{ |c| }{MLP} \\
\hline
1x1x1 conv($C_{\text{output}}$), GroupNorm(8), Swish & $D\times D\times D\times C_{\text{output}}$ & 1  \\

\hline
\hline
\multicolumn{3}{ |c| }{Sampling \& Grouping} \\
\hline
\multicolumn{3}{ |c| }{Number of Centers: $N_\text{center}$}\\
\hline
\multicolumn{3}{ |c| }{Grouping Radius: $r$}\\
\hline
\multicolumn{3}{ |c| }{Number of Neighbors: $N_\text{neighbor}$}\\
\hline
\end{tabular}
\caption{Set Abstraction Layer. Input is first fed through $L$ PVConv modules, then to an MLP module, and finally through the Sampling \& Grouping module.}
\label{tab:sa}
\end{table*}

\begin{table*}[h!]
\centering
\begin{tabular}{ |c|c|c| }
%\label{celeba}
\hline
\multicolumn{3}{|c|}{\bf{Feature Propagation}}\\
\hline
\multicolumn{3}{ |c| }{Input Feature Size: $N_{\text{input}} \times C_{\text{input}}$} \\
\hline
\multicolumn{3}{ |c| }{Output Feature Size: $N_{\text{output}} \times C_{\text{output}}$} \\
\hline
\multicolumn{3}{ |c| }{Voxelization Resolution: $D$}\\
\hline
\multicolumn{3}{ |c| }{Number of Point-Voxel Convolution (PVConv) blocks: $L$}\\
\hline
\multicolumn{3}{ |c| }{Whether to use attention mechanism: use\_attn}\\
\hline
\hline
\multicolumn{3}{ |c| }{Interpolation} \\
\hline

\hline
\multicolumn{3}{ |c| }{PVConv $\times L$} \\
\hline
Layers & In-Out Size & Stride \\ \hline
Input: X & $D\times D\times D\times C_{\text{input}}$ & \\
3x3x3 conv($C_{\text{output}}$), GroupNorm(8), Swish & $D\times D\times D\times C_{\text{output}}$ & 1 \\
Dropout(0.1) & &\\
3x3x3 conv($C_{\text{output}}$), GroupNorm(8) & $D\times D\times D\times C_{\text{output}}$ & 1  \\
Attention(use\_attn) & $D\times D\times D\times C_{\text{output}}$ & \\
\hline
\hline
\multicolumn{3}{ |c| }{MLP} \\
\hline
3x3x3 conv($C_{\text{output}}$), GroupNorm(8), Swish & $D\times D\times D\times C_{\text{output}}$ & 1  \\

\hline
\end{tabular}
\caption{Feature Propagation Layer. Input is fed through Interpolation module, $L$ PVConv modules, and an MLP module.}
\label{tab:fp}
\end{table*}

\begin{table*}[h!]
\centering
\begin{tabular}{ |c|c|c|c|c|}
\hline
\multicolumn{5}{ |c| }{Input Feature Size: $2048 \times 3$} \\
\hline
\multicolumn{5}{ |c| }{Input Time Embedding Size: 64} \\
\hline

\multicolumn{5}{ |c| }{Output Feature Size: $2048 \times 3$} \\
\hline
\hline
\multicolumn{5}{ |c| }{Time Embedding} \\
\hline
\multicolumn{5}{|c|}{Sinusoidal Embedding dim = 64} \\
\multicolumn{5}{|c|}{MLP(64, 64)} \\ 
\multicolumn{5}{|c|}{LeakyRelU(0.1)} \\ 
\multicolumn{5}{|c|}{MLP(64, 64)} \\ 

\hline
\hline

& SA 1 & SA 2 & SA 3 & SA 4 \\
\hline
$L$ & 2&3&3&0\\
$C_{\text{input}}$ & 3&32&64& - \\
$E_t$ & 64&64&64& - \\
$C_{\text{output}}$ & 32&64&128& - \\
$D$ & 32 & 16&8&-\\
use\_attn & False &True &False &- \\
$N_\text{center}$ & 1024 &256&64&16 \\
$r$ & 0.1&0.2&0.4&0.8\\
$N_\text{neighbor}$& 32 &32&32&32\\
\hline
\hline
& FP 1 & FP 2 & FP 3 & FP 4 \\
\hline
$L$ & 3&3&2&2\\
$C_{\text{input}}$ & 128 &256&256&128 \\
$C_{\text{output}}$ & 256&256&128&64 \\
$D$ & 8&8&16&32\\
use\_attn & False&True&False&False \\
\hline
\hline

\multicolumn{5}{ |c| }{MLP(64,3)} \\
\hline
\end{tabular}
\caption{Entire point-voxel CNN architecture. Input point clouds and time steps are sequentially passed through SA 1-4, FP 1-4, and an MLP to obtain output of the same dimension. At the start of each SA and FP module, time embedding and point features are first concatenated.}
\label{tab:pvcnn}
\end{table*}

\subsection{Choices of $\beta_t$ and $T$}

For both hyper-parameters, we follow \cite{ho2020denoising}. For Car and Chair, we set $\beta_0=10^{-4}, \beta_T=0.01$ and linearly interpolate other $\beta$'s. For Airplane, we interpolate between $\beta_0=10^{-5}$ and $\beta_T=0.008$ for the first 90\% steps and then fix $\beta_T=0.008$. We also set $T=1000$ for all experiments and we generally notice that lower timesteps (\eg, 100) are not enough for the model to construct shapes.

\subsection{Training Parameters}
We use Adam optimizer with learning rate $2\times 10^{-4}$ for all experiments.

\end{document}